%% file: main.tex
\documentclass[10pt,twocolumn,letterpaper]{article}

\usepackage{cvpr}
\usepackage{hyperref}
\usepackage{url}
\usepackage[svgnames]{xcolor}
\usepackage{tcolorbox}
\tcbuselibrary{skins, breakable}
\usepackage{multirow}
\usepackage{tabularx}
\usepackage{array}
\usepackage{subcaption}
\usepackage{wrapfig}
\usepackage{xcolor}
\usepackage[utf8]{inputenc} 
\usepackage[T1]{fontenc}    
\usepackage{url}            
\usepackage{booktabs}       
\usepackage{amsfonts}       
\usepackage{nicefrac}       
\usepackage[utf8]{inputenc}
\usepackage[T1]{fontenc}
\usepackage{xcolor}

\usepackage{pifont}
\usepackage{xcolor}
\definecolor{codegreen}{rgb}{0,0.6,0}
\definecolor{codegray}{rgb}{0.5,0.5,0.5}
\definecolor{codepurple}{rgb}{0.58,0,0.82}
\definecolor{backcolour}{rgb}{0.95,0.95,0.92}
\usepackage{algorithm}
\usepackage{algpseudocode}
\renewcommand{\algorithmicrequire}{\textbf{Input:}}

\usepackage{listings}
\usepackage{multirow} 
\usepackage{makecell} 
\usepackage{caption} 
\usepackage{graphicx}   
\usepackage{colortbl}   
\usepackage{tabularx}
\usepackage{tcolorbox}
\tcbuselibrary{skins, breakable}
\usepackage{microtype}      
\usepackage{xcolor}         
\usepackage{caption}
\usepackage{float}
\usepackage{url}            
\usepackage{booktabs}       
\usepackage{amsfonts}       
\usepackage{nicefrac}       
\usepackage{microtype}      
\usepackage{lipsum}		
\usepackage{graphicx}
\usepackage{natbib}
\usepackage{doi}
\usepackage{times}
\usepackage{multirow}
\usepackage[table]{xcolor} 
\usepackage{soul}
\usepackage{bbding}
\usepackage{url}
\usepackage{amsmath}
\usepackage{amsthm}
\usepackage{algorithm}
\usepackage{newfloat}
\usepackage{bbding}
\usepackage{listings}
\usepackage{tikz}
\usepackage{comment}
\usepackage{amssymb}
\usepackage{color}
\usepackage[utf8]{inputenc}
\usepackage[T1]{fontenc}
\usepackage{xcolor}

\renewcommand{\algorithmicrequire}{\textbf{Input:}}

\usepackage{hyperref}
\hypersetup{
    colorlinks=true,
    citebordercolor=blue,
    linkbordercolor=red,
    urlbordercolor=magenta
}

\definecolor{cvprblue}{rgb}{0.21,0.49,0.74}
\def\paperID{xxxx}
\def\confName{CVPR}
\def\confYear{2026}

\title{Light-Omni: Reflex over Reasoning in Agentic Video Understanding with Long-Term Memory}
\author{
    Chang Nie \quad
    Jiaju Wei \quad 
    Junlan Feng\thanks{Corresponding author.} \quad
    Chaoyou Fu \quad
    Caifeng Shan$^{\spadesuit}$ \vspace{2mm}\\
    Nanjing University  \\
    {\tt\small changnie@smail.nju.edu.cn}
}
\makeatletter
\def\@fnsymbol#1{\ensuremath{\ifcase#1\or \spadesuit \or \dagger \or \ddagger \or \mathsection \or \mathparagraph \or \| \or ** \or \dagger\dagger \or \ddagger\ddagger \else\@ctrerr\fi}}
\makeatother

\begin{document}
\maketitle
\input{0_abstract}

\input{1_intro}

\input{2_rele}

\input{3_method}

\input{5_exp}

{
    \small
    \bibliographystyle{ieeenat_fullname}
    \bibliography{main}
}
\input{X_suppl}

\end{document}

%% file: 0_abstract.tex
\begin{abstract}
Agentic video understanding equips models with long-term memory to autonomously process and respond to continuous, long-horizon multimodal streams.
However, advanced video agents often rely on ``detective-style'' iterative reasoning for action control (e.g., $\mathtt{search}$) and evidence aggregation, incurring prohibitive costs and latency.
We argue that such heavy reasoning primarily compensates for the lack of global context and semantic misalignment in retrieval.
This paper introduces \textbf{\textit{Light-Omni}}, a multimodal agent framework for reflexive and lightweight video understanding.
It achieves this through dual contextual states that instantly build the required context in a single forward pass.
First, we maintain a \textit{\textbf{global state}}, a finite-sized multimodal script continuously consolidated from episodic memory, serving as the global context for Light-Omni.
Through hierarchical merging, it preserves recent details while summarizing past events.
Second, conditioned on this global context, we generate a parametric \textbf{\textit{latent state}} that directly drives autonomous actions and produces retrieval embeddings, with minimal latency.
Benefiting from this coupled design, Light-Omni achieves semantically aligned retrieval and reflexive responses while avoiding iterative reasoning.
Extensive experiments validate the effectiveness of Light-Omni across multiple video benchmarks.
Notably, it outperforms M3-Agent with an average \textbf{2.4}\% accuracy gain, a \textbf{12.1}$\times$ speedup, and a \textbf{2.6}$\times$ improvement in GPU memory efficiency.
Furthermore, it serves as a memory system to enhance both the performance and efficiency of existing MLLMs.
Project page: {\hypersetup{hidelinks}\href{https://clare-nie.github.io/Light-Omni/}{\color[rgb]{0.8, 0.2, 0.5} https://clare-nie.github.io/Light-Omni/}}.
\end{abstract}

%% file: 1_intro.tex
\section{Introduction}
\label{sec:intro}

\begin{figure*}[t]
\centering
\includegraphics[width=.99\textwidth]{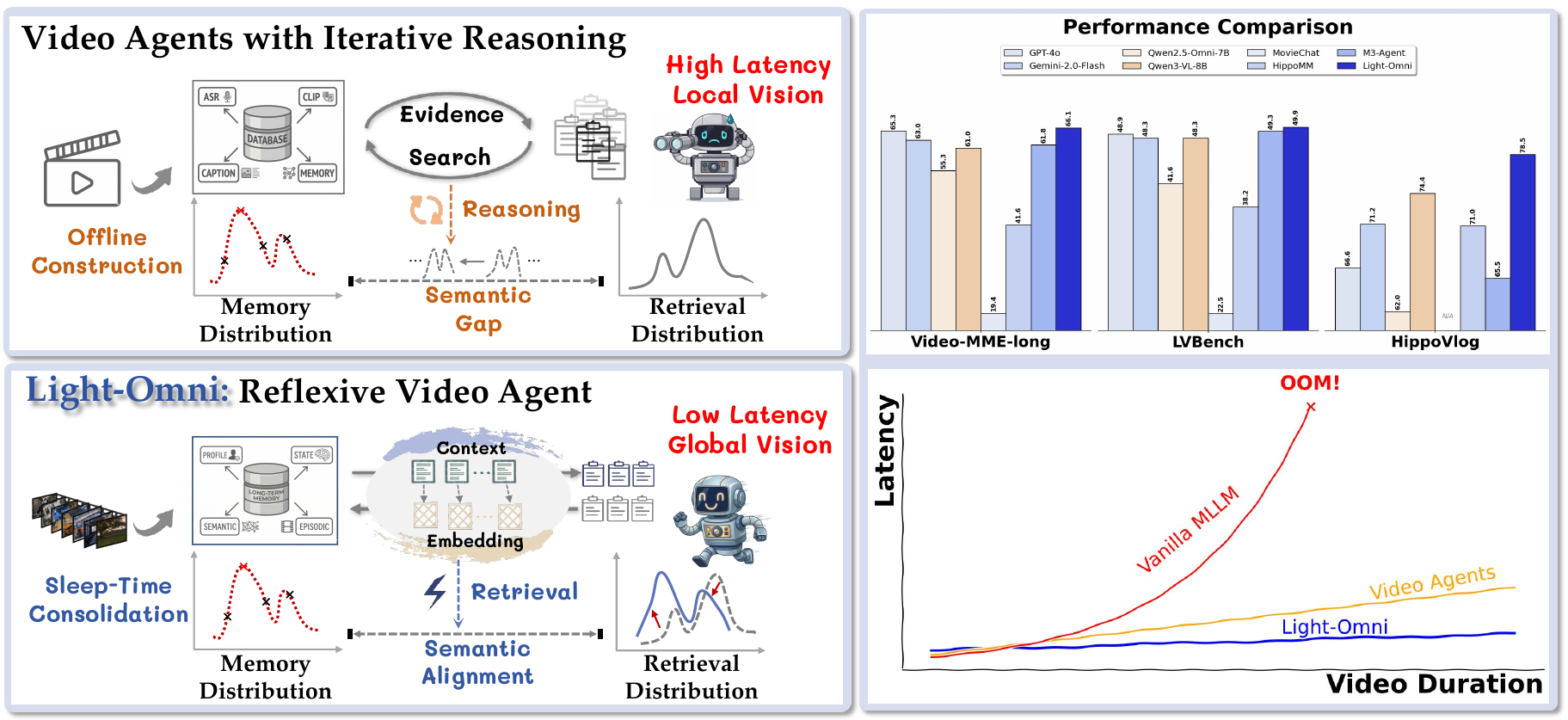} 
\caption{
\textbf{Comparison of video understanding paradigms and overall performance.} \textbf{(Left)} Recent video agents rely on ``detective-style'' iterative reasoning to bridge the semantic gap between query and offline-constructed memory representations, suffering from high latency and local vision.
In contrast, \textbf{Light-Omni} leverages dual contextual states to maintain a global context and directly generate semantically aligned embeddings for evidence aggregation. By incorporating sleep-time memory consolidation\protect\footnotemark, it enables seamless online interaction with both global vision and near-constant latency.
\textbf{(Right)} Light-Omni achieves strong performance on long-video benchmarks while maintaining near-constant latency regardless of video duration.
\vspace{-2mm}}
\label{fig:intro}
\end{figure*}

It is a cornerstone of human cognition that our actions are guided by memory~\cite{gabrieli1998cognitive}.
Yet, in the film Memento, Leonard, suffering from anterograde amnesia, must rely on external data of Polaroid photos and notes to maintain a fleeting connection to his own past~\cite{gargett2002nolan}.
His struggle serves as a metaphor for current Multimodal Large Language Models (MLLMs) when confronted with continuous, long-horizon video streams~\cite{chen2024evolution,hatalis2023memory}.
Bound by finite context windows, these models are essentially amnesiacs.
To mitigate this inherent limitation, the paradigm of memory-augmented agentic video understanding~\cite{song2024moviechat,zhang2025deep,long2025seeing} has emerged, where systems sift through their own external ``Polaroids'' to recall past events.

In practice, these Polaroids correspond to external databases or evolving memory banks, accessed via retrieval mechanisms to compensate for missing context~\cite{zhao2026retrieval,hatalis2023memory,xu2025mem}.
Most existing approaches~\cite{zhong2024memorybank,xu2025mem,kang2025memory} rely on similarity search over pre-indexed memory entries.
In video understanding, this requires segmenting continuous streams into discrete clips, typically annotated with coarse textual metadata such as captions or summaries~\cite{jing2023memory,zhang2024internlm,tao2025omniagent,zhang2025deep}.
However, real-world video queries are inherently context-dependent and often contain noise, coreference, and implicit cues.
As a result, a substantial semantic gap arises between user queries and stored memory representations, degrading retrieval accuracy and introducing significant redundancy~\cite{gao2023precise,zhou2024trustworthiness}.
To mitigate this issue, recent methods explore LLM-driven retrieval optimizations, including query rewriting~\cite{zhao2024retrieval,ma2023query}, condition generation~\cite{nie2026}, and agentic memory construction~\cite{kang2025memory,long2025seeing}.
For video agents~\cite{zhang2025deep,long2025seeing,yeo2025worldmm,zhang2025thinking}, this has evolved into a ``detective-style'' paradigm involving iterative reasoning, planning, and retrieval.
This process mirrors Leonard’s struggle to cross-reference clues and piece together the truth.
While effective, such workflows introduce significant cost and latency~\cite{zhang2025deep}.

{We argue that the reliance on heavy reasoning mainly stems from the lack of an explicit global context and the misalignment between the query and memory distributions.}
Specifically, existing memory systems fail to model a \textit{global context}, i.e., a holistic and persistent representation of past observations.
Instead, they rely on fragmented substitutes that still suffer from the very context loss they aim to resolve.
For instance, short-term memory~\cite{hu2025memory} and sparse sampling~\cite{wang2024videoagent,tang2025adaptive} break temporal coherence, while efficiency-oriented techniques such as KV-caching optimization~\cite{xiao2023efficient,kim2025infinipot} and token merging~\cite{song2024moviechat} struggle to scale to long-horizon inputs.
Accordingly, without a global context, retrieval becomes a \textit{myopic hunt} that prioritizes local similarity while ignoring global narrative structure.

Fortunately, the normal human mind is spared such profound context loss, innately forming a cognitive map to guide attention~\cite{lin2025hippomm,kahneman2011thinking}.
This global context catalyzes reflexive action over deliberate
reasoning—allowing memory to be evoked by context, rather
than merely searched.
Motivated by this, we propose \textbf{Light-Omni}, a multimodal agent framework that transitions video understanding from multi-step reasoning to reflexive response generation (see Fig.~\ref{fig:intro}).
We first build a multimodal long-term memory system composed of identity profiles, alongside semantic and episodic components.
Building upon this foundation, we construct dual contextual states:
(1) \textbf{\textit{Global State}}: a non-parametric and compact multimodal script consolidated from episodic memory at sleep-time. Through resolution-decaying hierarchical merging, it preserves both recent observations ($\mathtt{visual}$, $\mathtt{auditory}$, $\mathtt{interactive}$) and long-range context;
(2) \textbf{\textit{Latent State}}: a parametric representation that enables reflexive action control via task-specific heads (e.g., $\mathtt{speech}$, $\mathtt{search}$) and directly provides semantically aligned embeddings for retrieval.
This design bridges the semantic gap and enables precise retrieval without reasoning.

Our method is driven by two insights.
First, human memory exhibits temporal decay, which has been widely adopted in memory system design~\cite{zhong2024memorybank,kang2025memory}.
Second, we observe that explicit retrieval intermediaries (e.g., $\mathtt{rewrites}$, $\mathtt{conditions}$, and $\mathtt{keywords}$) introduce unnecessary bottlenecks that hinder alignment between query and memory distributions.
Inspired by~\cite{hao2024training,wang2025m+}, we directly learn retrieval embeddings by jointly optimizing the backbone and the embedding space.
In summary, our main contributions are:
\footnotetext{An asynchronous process that continuously updates memory without blocking real-time interactions.}
\begin{itemize}
\item We propose \textbf{Light-Omni}, a multimodal agent framework for reflexive and lightweight video understanding with long-term memory. It serves as a general memory system that can be seamlessly integrated into existing MLLMs to improve both efficiency and performance.

\item We introduce dual contextual states to bridge the semantic gap between query and memory distributions, enabling accurate and robust retrieval under noisy inputs without the burden of iterative reasoning.

\item Light-Omni achieves strong gains over both the baseline (Qwen2.5-Omni-7B) and the strong counterpart M3-Agent with \textbf{9.5\% and 2.4\% accuracy} improvements, \textbf{20.5$\times$} and \textbf{12.1$\times$} \textbf{speedups}, and \textbf{3.3$\times$} and \textbf{2.6$\times$ }GPU memory reductions, respectively.
\end{itemize}

%% file: 2_rele.tex
\section{Related Work}
\label{sec:rel}

\noindent\textbf{Omni-Modal Models.}
Recent advancements in MLLMs mark a paradigm shift, transitioning from vision-centric designs to unified, omni-modal architectures~\cite{hurst2024gpt,wu2024next}.
The Qwen-Omni series~\cite{xu2025qwen2,Qwen3Omni} and the Gemini series~\cite{team2024gemini,comanici2025gemini},
representing both open-source and proprietary frontiers, exhibit sophisticated multimodal understanding, capable of processing interleaved streams of video, audio, and text in an end-to-end manner.
However, when confronted with long-horizon streaming inputs, these models remain constrained by finite context windows and prohibitive computational complexity.
Prevailing methods rely on sparse sampling or efficient attention strategies~\cite{tang2025adaptive,xiao2023efficient,zhang2025stream}, which inevitably sacrifice fine-grained details or temporal coherence.
As a result, while providing a strong perceptual foundation, these models necessitate memory-augmented architectures to preserve temporal continuity across long-horizon interactions.

\vspace{2mm}
\noindent\textbf{Retrieval- and Memory-Augmented Video Understanding.}
To address the limitations of finite context windows, recent works introduce retrieval-augmented paradigms for long-horizon video understanding~\cite{xu2025mem,long2025seeing,xu2026long}.
Early approaches maintain static memory banks composed of compressed frame features or sparse textual metadata, such as clip summaries and ASR transcripts~\cite{wang2024videoagent,luo2024video}, following retrieval-augmented generation (RAG) paradigms~\cite{zhao2024retrieval}.
To preserve temporal structure and event continuity, more recent methods further organize memory into semantic and episodic components, or adopt agent-driven strategies for dynamic memory construction~\cite{hu2025memory}.
Despite these advances, retrieval remains a fundamental bottleneck in video understanding.
Most existing methods rely on shallow semantic matching between user queries and stored memory entries, which is particularly fragile under noisy, context-dependent, or implicit queries.
This limitation stems from a persistent distribution gap between user queries and sparse memory representations, leading to misalignment and redundant retrieval results~\cite{zhou2024trustworthiness,liu2024bridging,chan2024rq,gupta2024comprehensive}.

\vspace{2mm}
\noindent\textbf{Agentic Video Understanding.}
To further improve retrieval quality and task performance, recent video agents have shifted from passive retrieval toward active, goal-oriented reasoning and execution~\cite{long2025seeing,yeo2025worldmm,tao2025omniagent}.
These methods typically adopt a ``detective-style'' workflow, where multi-turn reasoning is used to decompose queries, refine search intents, and iteratively invoke external tools (e.g., $\mathtt{ASR}$, $\mathtt{search}$, $\mathtt{grounding}$) for evidence aggregation.
Representative systems such as OmniAgent~\cite{tao2025omniagent} and LongVideoAgent~\cite{liu2025longvideoagent} further incorporate hierarchical planning or modality-aware tool selection to improve robustness in complex video scenarios.
While these approaches improve retrieval accuracy and downstream performance, they introduce substantial computational overhead due to repeated reasoning and tool invocation.
Such multi-step interaction also significantly increases latency, making highly responsive agentic interactions 
difficult to achieve in practice~\cite{long2025seeing,zhang2025deep}.
More fundamentally, this paradigm still treats reasoning as a compensatory mechanism for imperfect retrieval, rather than directly addressing the underlying misalignment between query semantics and memory representations.

%% file: 3_method.tex
\section{Light-Omni Framework}
\label{sec:method}

\begin{figure*}[t]
\centering
\includegraphics[width=.99\textwidth]{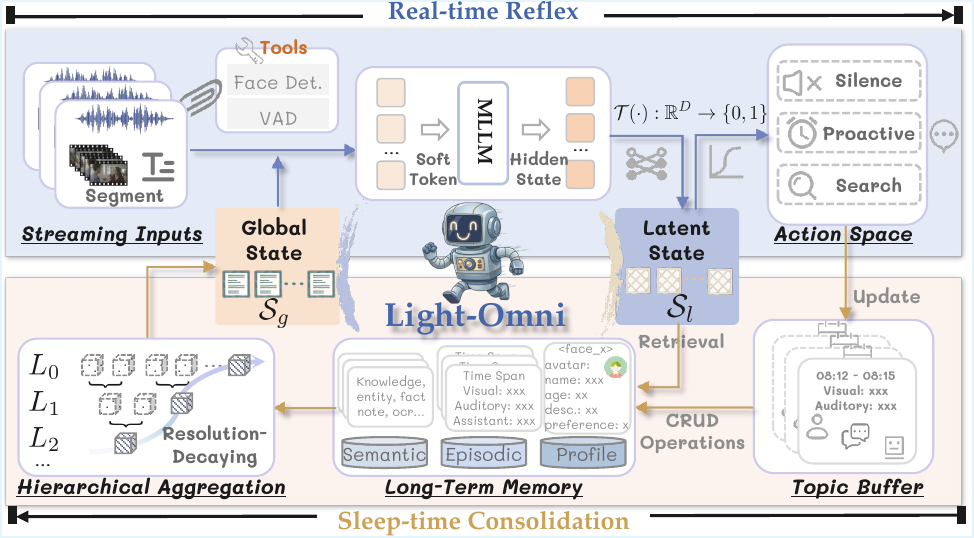} 
\caption{\textbf{Overall architecture of Light-Omni.} It operates via dual contextual states augmented with long-term memory.
\textbf{(Top)} \textit{Real-time Reflex} (\protect\scalebox{1.2}{\textcolor[HTML]{2B31CE}{$\boldsymbol{\rightarrow}$}}): Light-Omni continuously processes streaming inputs to generate the latent state $\boldsymbol{S_l}$ for instantaneous action control and context-aware retrieval. \textbf{(Bottom)} \textit{Sleep-time Consolidation} (\protect\scalebox{1.2}{\textcolor[HTML]{b97D0E}{$\boldsymbol{\rightarrow}$}}): The system asynchronously manages long-term memory via CRUD (Create, Read, Update, Delete) operations and hierarchical aggregation to refresh the global state $\boldsymbol{S_g}$. This coupled design ensures reflexive and lightweight video understanding.\vspace{-5mm}}
\label{fig:fra}
\end{figure*}

\noindent\textbf{Overview.}
The overall architecture of \textbf{Light-Omni} is illustrated in Fig.~\ref{fig:fra}.
It operates via a dual contextual state design, decoupling the slow, incremental consolidation of global context from the fast, real-time reflexive response driven by the latent state.
Given an omni-modal input segment at timestamp $t$, $\boldsymbol{\mathcal{I}}^t=\{\boldsymbol{V}^t, \boldsymbol{A}^t, \boldsymbol{T}^t\}$, comprising visual frames, audio signals, and textual instructions, the model produces actions conditioned on the current context while updating memory in a post-response consolidation phase.
This process is mathematically formalized as follows:
\begin{equation}
\{\boldsymbol{\mathcal{A}}^t, \boldsymbol{\mathcal{S}}^t, \boldsymbol{\mathcal{M}}^t\} = \Phi(\pi\ ;\ \boldsymbol{\mathcal{I}}^t, \boldsymbol{\mathcal{S}}^{t-1}, \boldsymbol{\mathcal{M}}^{t-1}), \quad t=0, 1, 2, \cdots
\label{eq1}
\end{equation}
where $\boldsymbol{\mathcal{A}}^t$ denotes reflexive actions (e.g., response generation, search), $\boldsymbol{\mathcal{S}}^t$ represents the dual contextual states, and $\boldsymbol{\mathcal{M}}^t$ is the structured multimodal long-term memory. $\Phi(\cdot)$ defines the system dynamics, parameterized by the model $\pi$ (e.g., Qwen2.5-Omni-7B~\cite{xu2025qwen2}).

Light-Omni aims to equip general MLLMs with long-term memory and low-latency inference capabilities for long video understanding.
While recent video agents~\cite{song2024moviechat,long2025seeing,zhang2025deep,tao2025omniagent} achieve strong performance, they rely on a ``detective-style'' workflow involving iterative reasoning and tool invocation for evidence aggregation. 
Such multi-step deliberation introduces prohibitive latency, high memory overhead, and dependence on external APIs. 
In contrast, our framework achieves near-constant latency by enabling reflexive action control and semantically aligned retrieval within a single forward pass.

\subsection{Multimodal Long-Term Memory}
\label{sec:method:memory}
Memory serves as the cornerstone of Light-Omni, enabling the long-term behavioral consistency essential for agentic systems~\cite{hu2025memory}.
Before detailing the dual-state workflow, we formalize the construction of the multimodal memory system $\boldsymbol{\mathcal{M}}$.
The primary objectives of this memory system are twofold: (1) storing \textit{omni-modal} streams alongside continuous user-assistant interactions, and (2) supporting structured CRUD operations for dynamic knowledge maintenance and highly efficient retrieval.
Inspired by recent architectures~\cite{long2025seeing,xu2025mem}, we decompose the multimodal memory $\boldsymbol{\mathcal{M}}$ into three primary and distinct components:

\begin{itemize}
    \item \textbf{User Profile $\boldsymbol{\mathcal{M}}_p$} stores fundamental attributes of individuals, including visual avatars, personal preferences, and personality traits. This empowers the agent to recognize users across time and deliver personalized interactions.

    \item \textbf{Semantic Memory $\boldsymbol{\mathcal{M}}_s$} distills abstracted facts, concepts, and relationships from streaming inputs. Stored as textual key-value pairs (timestamps and content), it allows the agent to quickly recall crucial knowledge without sifting through granular historical logs.

    \item \textbf{Episodic Memory $\boldsymbol{\mathcal{M}}_e$} serves as a chronological ledger of past events. It logs historical interactions as structured, omni-modal scripts comprising explicit timestamps, visual scene descriptions, auditory cues, and assistant responses.
\end{itemize}

For efficiency, Light-Omni processes the continuous video stream sequentially. 
Given the current input $\boldsymbol{\mathcal{I}}^t$ and the global state $\boldsymbol{S}_g^{t-1}$, we directly generate $\text{Topic}^t = \{\boldsymbol{\mathcal{M}}_s^t, \boldsymbol{\mathcal{M}}_e^t\}$ via $p(\text{Topic}^t \mid \boldsymbol{\mathcal{I}}^t, \boldsymbol{S}_g^{t-1}; \pi)$.
It captures both semantic facts and episodic events of the current segment.
The user profile $\boldsymbol{\mathcal{M}}_p$ is updated only when the number of buffered topics exceeds a predefined capacity $\tau_t$.
Crucially, for offline video understanding (e.g., benchmark evaluation), this memory construction is executed in advance, whereas in online scenarios, the entire process operates asynchronously during the system's \textit{sleep-time}, ensuring seamless, real-time preservation of multimodal memories without blocking active user interactions.

Light-Omni offers three key advantages over existing memory systems~\cite{yeo2025worldmm,song2024moviechat}:
(i) \textbf{Narrative-Style:} Episodic memory is incrementally generated, guaranteeing narrative continuity. 
(ii) \textbf{Human-Centric:} It centers on the user, establishing connections across different memory types via specific identifiers (e.g., \texttt{<face\_id>})~\cite{long2025seeing}.
(iii) \textbf{Lightweight \& Omni-Modal:} It distills audio-visual and textual streams into compact representations with minimal computational overhead.
Further details on memory architecture and its respective functions are provided in the supplemental material.

\subsection{Dual-State Design}
At the core of Light-Omni lies a novel dual-state design, $\boldsymbol{\mathcal{S}}^t = \{\boldsymbol{S}_g^t, \boldsymbol{S}_l^t\}$, where both states are derived from and grounded in the multimodal memory $\boldsymbol{\mathcal{M}}^{t-1}$.
This mechanism ensures that: (i) $\boldsymbol{S}_g^t$ endows the agent with a global context for response generation, memory retrieval, and consolidation; and (ii) $\boldsymbol{S}_l^t$ enables action execution via fast, implicit \textit{reflexes} rather than heavy, explicit \textit{reasoning}.
This dual-state design allows Light-Omni to bridge the semantic gap between queries and memory representations through coordinated global context and semantically-aligned retrieval.

\vspace{2mm}
\noindent\textbf{Global State ($\boldsymbol{S}_g$):} 
Acting as a non-parametric contextual backbone, the global state $\boldsymbol{S}_g$ provides a compact, hierarchical representation of historical context.
Rather than naively concatenating the entire memory bank, $\boldsymbol{S}_g$ is constructed via a hierarchical merging strategy that exhibits a \textit{resolution-decaying} property.
Specifically, the episodic scripts $\boldsymbol{\mathcal{M}}_e$ are progressively summarized and merged based on a capacity factor $k$.
When the number of level-$i$ nodes $N_i$ reaches $k+1$\footnote{To prevent cliff forgetting, merging occurs only when $N_i \geq k+1$, ensuring the most recent node at each level remains preserved. We empirically set $k=8$ to trade-off performance and overhead.}, the oldest $k$ nodes are consolidated into a single higher-level node $\boldsymbol{m}_{i+1}$:
\begin{align}
    \boldsymbol{S}_g^{t} &= \mathit{Update}(\boldsymbol{\mathcal{M}}_e^{t-1} \cup \Delta\boldsymbol{\mathcal{M}}_e^t, k) \label{eq:global_update} \\
    \text{s.t.} \quad & N_i \geq k+1 \implies \boldsymbol{m}_{i+1} \leftarrow \mathit{Merge}(\{\boldsymbol{m}_i\}_{1:k})\notag.
\end{align}
This operation ensures that recent fine-grained details and distant high-level summaries coexist within a bounded context window, preserving temporal continuity and coherence.

\vspace{2mm}
\noindent\textbf{Latent State ($\boldsymbol{S}_l$):}
Conditioned on $\boldsymbol{S}_g^{t}$, it directly generates semantically aligned embeddings for accurate retrieval.
Inspired by~\cite{hao2024training,wang2025m+}, rather than explicitly generating textual thoughts, we introduce learnable soft prompts $\boldsymbol{P}_{soft}\in\mathbb{R}^{N\times D}$ appended to the input sequence.
Within a single forward pass, the backbone model $\pi_\theta$ processes the augmented sequence $[\boldsymbol{S}_g^{t}, \boldsymbol{\mathcal{I}}^t, \boldsymbol{P}_{soft}]$ and directly yields corresponding hidden features $\boldsymbol{H}^t$. These specific hidden states are then decoded in parallel via task-specific heads:
\begin{equation}
\begin{aligned}
\boldsymbol{z}_{ret}^t &= \text{Pro}(\boldsymbol{h}_{ret}^t) \in \mathbb{R}^{D'}, && (\text{Retrieval Embedding}) \\
\boldsymbol{a}_{act}^t &\sim \mathrm{Bernoulli}\big( \boldsymbol{p}_{act}^t \big) \in \{0, 1\}^2, && (\text{e.g., } \mathtt{speech}, \mathtt{search})
\end{aligned}
\label{eq:latent_generation}
\end{equation}
\noindent where $\boldsymbol{p}_{act}^t = \text{Sigmoid}(\text{FC}(\boldsymbol{h}_{act}^t))$ represents the independent trigger probability for each action. Here, $\boldsymbol{h}_{act}^t$ and $\boldsymbol{h}_{ret}^t$ are distinct hidden state vectors extracted from $\boldsymbol{H}^t$, corresponding to the soft prompt tokens.
The projection head $\text{Pro}(\cdot)$ is an \textit{MLP-ReLU-MLP} module, mapping the hidden state into embedding space of dimension $D'$.
Then, $\boldsymbol{z}_{ret}^t$ is added to the original retrieval embedding to \textit{rectify} the query representation to match the memory distribution, as illustrated in Fig.~\ref{fig:intro}.
By natively rectifying distribution discrepancies in the continuous latent space, Light-Omni concurrently evaluates \textit{whether} to act and \textit{what} to retrieve.

\subsection{Learning Strategy and Optimization}
\label{sec:method:opt}

\noindent\textbf{Dataset Construction.} 
We develop an automated pipeline to synthesize a customized training dataset from publicly available offline long videos.
Specifically, we inject textual and auditory instructions at diverse time steps to generate synchronized responses and dense intermediate supervisory signals (e.g., memory updates and executed actions).
To enhance diversity and simulate multi-session dynamics, heterogeneous video clips are concatenated.
This yields 43$k$ training samples. Further details are provided in the supplemental material.

\vspace{2mm}
\noindent\textbf{Multi-LoRA Design.}
The core capabilities of Light-Omni are categorized into three dimensions: memorization, generation, and reaction. To mitigate optimization conflicts (e.g., interference between memory retrieval and text generation), we employ a multi-LoRA~\cite{hu2022lora} architecture to decouple these tasks. During inference, the system dynamically switches adapters based on the current state.

\vspace{2mm}
\noindent\textbf{Training.}
For memorization and generation tasks, we optimize the corresponding adapters using the standard Next Token Prediction (NTP~\cite{brown2020language}) objective: $\mathcal{L}_1 = -\sum_{t=1}^{T} \log P_\theta(y_t \mid x, y_{<t})$.
In contrast, the reaction adapter is optimized via a hybrid objective combining discrete classification and contrastive retrieval alignment:
\begin{equation}
\mathcal{L}_{2} = \sum_{c \in \mathcal{C}} \mathcal{L}_{CE}(\boldsymbol{p}_{c}, \boldsymbol{y}_c) - \lambda \log \frac{\exp(\boldsymbol{z}_{ret} \cdot \boldsymbol{k}^+ / \tau)}{\sum_{j=1}^B \exp(\boldsymbol{z}_{ret} \cdot \boldsymbol{k}_j / \tau)},
\label{eq:loss}
\end{equation}
where $\mathcal{C}=\{\mathtt{search}, \mathtt{speech}\}$ denotes the action space\footnote{Light-Omni currently supports actions including $\mathtt{search}$ and $\mathtt{speech}$; additional capabilities (e.g., tool use) can be integrated with negligible cost.}, $\boldsymbol{z}_{ret}$ is the predicted retrieval embedding, and $\boldsymbol{k}^+$ is the corresponding positive memory embedding. Crucially, this formulation enables the \textit{joint} optimization of the backbone model and the latent retrieval representation.
\begin{figure}[t]
\centering
\includegraphics[width=.5\textwidth]{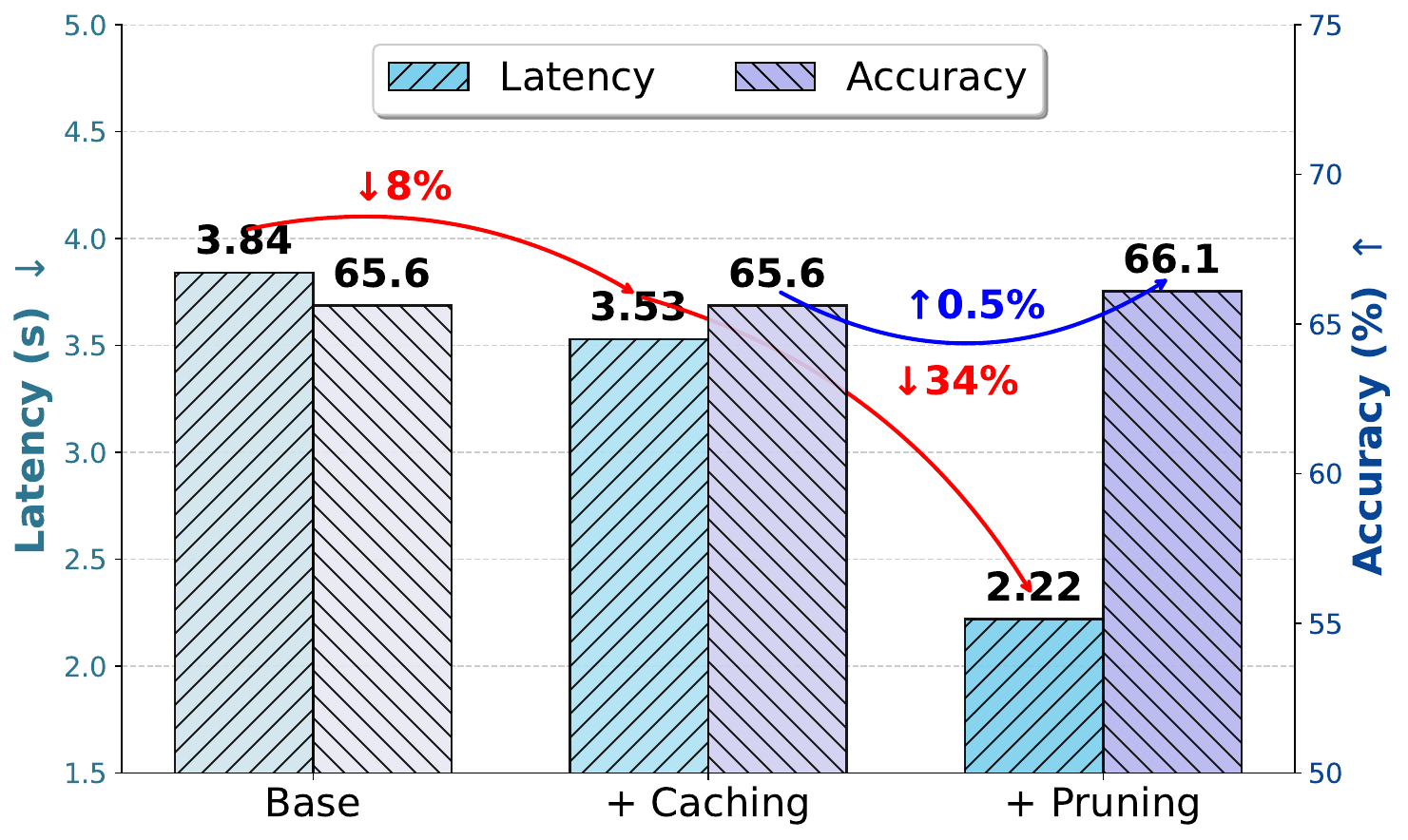} 
\caption{
Impact of efficiency optimization strategies. Feature caching and redundancy pruning cumulatively reduce inference latency by 42\% without compromising accuracy\protect\footnotemark.}
\label{fig:latency}
\end{figure}
\footnotetext{We evaluate the response latency over a ten-minute video, while assessing the performance on the VideoMME-long benchmark~\cite{fu2025video}.}

\begin{table*}[!h]
\centering
\caption{Performance and efficiency comparison on long-video benchmarks. We report accuracy (\%), latency (s), and memory footprint (GB).
Subscripts highlight the absolute accuracy gains and relative efficiency multiples (speedup/reduction) achieved by Light-Omni relative to their corresponding baseline models.
}
\label{tab:main_results}
\small
\setlength{\tabcolsep}{10.0pt}
\begin{tabular}{l ccc ccc c}
\toprule
\multirow{2}{*}{\textbf{Model}} & \multicolumn{3}{c}{\textbf{VideoMME-long~\cite{fu2025video}}} & \multicolumn{3}{c}{\textbf{LVBench~\cite{wang2025lvbench}}} & \multirow{2}{*}{\makecell{\textbf{Avg.} \\ \textbf{Acc. $\uparrow$}}} \\
\cmidrule(lr){2-4} \cmidrule(lr){5-7}
& Accuracy $\uparrow$ & Latency $\downarrow$ & Memory $\downarrow$ & Accuracy $\uparrow$ & Latency $\downarrow$ & Memory $\downarrow$ & \\
\midrule

\multicolumn{8}{l}{\textit{\textbf{Commercial \& Open-Source MLLMs}}} \\ \addlinespace[0.1cm]
GPT-4o~\cite{hurst2024gpt} & 65.3 & - & - & 30.8 & - & - & 48.1 \\
\rowcolor{blue!8}
Gemini-2.0-Flash~\cite{comanici2025gemini} & 63.0 & - & - & 48.6 & - & - & 55.8 \\
\rowcolor{blue!8}
Qwen2.5-Omni-7B~\cite{xu2025qwen2} & 55.3 & 48.8 & 81.7 & 41.6 & 51.9 & 80.0 & 48.5 \\
\rowcolor{blue!8}
Qwen2.5-VL-7B~\cite{bai2025qwen2} & 55.4 & 51.7 & 113.2 & 43.7 & 48.8 & 119.0 & 49.6 \\
Qwen2.5-VL-72B~\cite{bai2025qwen2} & 63.4 & - & OOM & 48.8 & - & OOM & 56.1 \\
Qwen3-VL-8B~\cite{yang2025qwen3} & 61.0 & 14.9 & 29.9 & 48.3 & 15.1 & 29.9 & 54.7 \\
\midrule

\multicolumn{8}{l}{\textit{\textbf{RAG \& Reasoning-based MLLMs}}} \\ \addlinespace[0.1cm]
Naive RAG  & 58.4 & 7.3 & 27.6 & 46.9 & 6.9 & 27.7 & 52.7 \\
RAG-Rewrite~\cite{ma2023query} & 60.8 & 8.5 & 27.7 & 47.6 & 7.4 & 27.9  & 54.2 \\
MovieChat~\cite{song2024moviechat} & 19.4 & - & 30.7 & 22.5 & - & 31.1 & 21.0 \\
HippoMM~\cite{lin2025hippomm} & 41.6 & 16.1 & 36.3 & 38.2 & 14.3 & 34.8 & 39.9 \\
M3-Agent~\cite{long2025seeing} & 61.8 & 25.5 & 62.4 & 49.3 & 33.1 & 62.2 & 55.6 \\
\midrule

\multicolumn{8}{l}{\textit{\textbf{MLLMs with Light-Omni (ours)}}} \\ \addlinespace[0.1cm]
\rowcolor{blue!8}
Qwen2.5-VL-7B~\cite{bai2025qwen2} & 59.6{\color{gray}$_{+4.2}$} & 8.5{\color{gray}$_{6.1\times}$} & 46.7{\color{gray}$_{2.4\times}$} & 49.3{\color{gray}$_{+5.6}$} &  6.8{\color{gray}$_{7.2\times}$} & 46.9{\color{gray}$_{2.5\times}$} & 54.5{\color{gray}$_{+4.9}$} \\
\rowcolor{blue!8}
Qwen3-VL-8B~\cite{yang2025qwen3} & 65.1{\color{gray}$_{+4.1}$} & 7.0{\color{gray}$_{2.1\times}$} & 46.3{\color{gray}$_{0.6\times}$} & 49.3{\color{gray}$_{+1.0}$} & 6.9{\color{gray}$_{2.2\times}$} & 46.9{\color{gray}$_{0.6\times}$} & 57.2{\color{gray}$_{+2.5}$} \\
\rowcolor{blue!8}
Gemini-2.0-Flash~\cite{comanici2025gemini} & 69.0{\color{gray}$_{+6.0}$} & - & - & 50.1{\color{gray}$_{+1.5}$} & - & - & 59.6{\color{gray}$_{+3.8}$} \\
\rowcolor{blue!20}

\midrule
\textbf{Light-Omni (ours)} & \textbf{66.1}{\color{gray}$_{+10.8}$} & \textbf{2.2}{\color{gray}$_{21.0\times}$} & \textbf{24.0}{\color{gray}$_{3.4\times}$} & \textbf{49.9}{\color{gray}$_{+8.3}$} & \textbf{2.6}{\color{gray}$_{20.1\times}$} & \textbf{24.2}{\color{gray}$_{3.3\times}$} & \textbf{58.0}{\color{gray}$_{+9.5}$} \\
\bottomrule
\end{tabular}
\end{table*}

\vspace{2mm}
\noindent\textbf{Efficiency Optimization.}
To further minimize latency, we implement two critical optimizations from preprocessing and workflow perspectives:
(1) \textbf{Feature Caching:} Since the omni-modal input at timestamp $t$ is utilized across multiple stages (e.g., response generation and memory update), we implement a caching mechanism to ensure that visual and auditory signals are encoded without repetition. This yields an initial 8\% reduction in latency.
(2) \textbf{Redundancy Pruning:} Leveraging the inherent temporal redundancy in long videos~\cite{alvar2025divprune}, we dynamically perform token pruning based on feature similarity.
As shown in Fig.~\ref{fig:latency}, these strategies cumulatively reduce inference latency by \textbf{42\%}.
Notably, this gain does not compromise performance; instead, it yields a slight accuracy improvement.

%% file: 5_exp.tex
\section{Experiments}
\label{sec:exp}

In this section, we conduct extensive experiments to validate the effectiveness and efficiency of \textbf{Light-Omni}. Our evaluation in the main paper is structured around four research questions:
\begin{itemize}
    \item \textbf{Question 1:} How does Light-Omni perform on long video benchmarks against strong baselines?
    \item \textbf{Question 2:} Can Light-Omni serve as a foundational memory system for existing MLLMs?
    \item \textbf{Question 3:} Does the dual-state mechanism guarantee retrieval precision and robustness under noisy queries?
    \item \textbf{Question 4:} How significantly does Light-Omni reduce latency and enhance user experience?
\end{itemize}

\noindent\textbf{Benchmarks \& Baselines.}
To rigorously evaluate the effectiveness of \textbf{Light-Omni}, we conduct extensive experiments on VideoMME-long~\cite{fu2025video}, LVBench~\cite{wang2025lvbench}, HippoVlog~\cite{lin2025hippomm}, and online OVO-Bench~\cite{niu2025ovo}.
These benchmarks assess the general capabilities of MLLMs on hour-long videos, including perception, audio-visual event understanding, and reasoning across diverse scenarios.
We compare Light-Omni against strong baselines including: \ding{172} general MLLMs, such as Qwen2.5-Omni-7B~\cite{xu2025qwen2}, Qwen2.5-VL-7B/72B~\cite{bai2025qwen2}, Qwen3-VL-8B~\cite{yang2025qwen3}, GPT-4o~\cite{hurst2024gpt}, and Gemini-2.0-Flash~\cite{comanici2025gemini}; \ding{173} memory-augmented methods like Naive RAG, RAG-Rewrite~\cite{ma2023query}, and MovieChat~\cite{song2024moviechat}; and \ding{174} reasoning-based agents like Ego-R1~\cite{tian2025ego}, HippoMM~\cite{lin2025hippomm}, M3-Agent~\cite{long2025seeing}, and WorldMM-8B~\cite{yeo2025worldmm}. This comprehensive comparison highlights the effectiveness of our method.

\begin{figure}[!t]
\centering
\includegraphics[width=.5\textwidth]{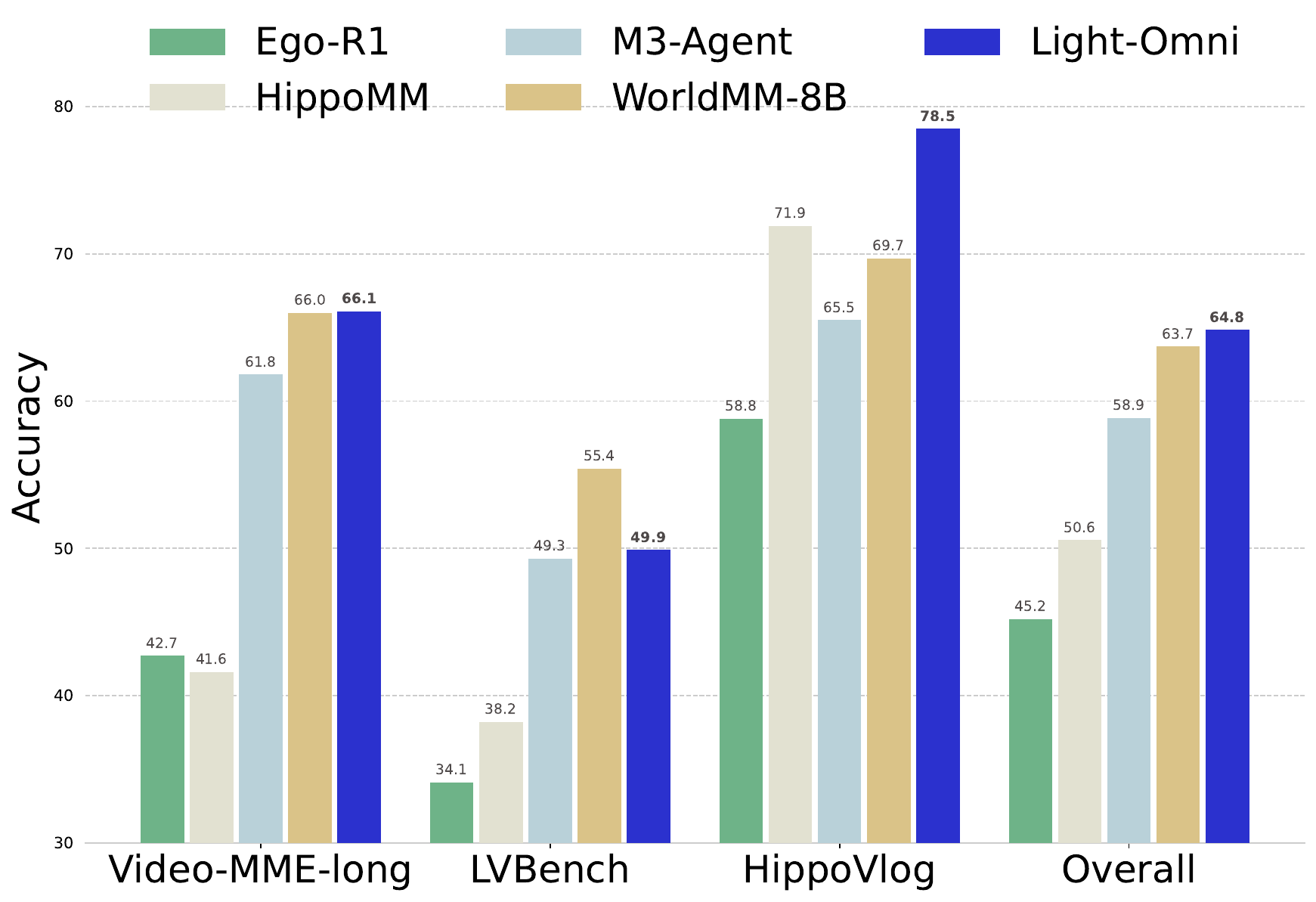} 
\caption{Comparison of Light-Omni and reasoning-based video agents. Light-Omni achieves the best overall performance across the benchmarks. Compared to the strong runner-up WorldMM-8B (which relies on GPT-5-mini for memory construction), Light-Omni not only yields a 1.1\% higher overall accuracy but also delivers an over 8$\times$ speedup (see Table~\ref{tab:efficiency}).
}
\label{fig:reasoning}
\end{figure}

\vspace{2mm}
\noindent\textbf{Implementation Details.}
In our experiments, we employ {Qwen2.5-Omni-7B}~\cite{xu2025qwen2} as the backbone model. The default dense retriever is Qwen3-Embedding-0.6B~\cite{zhang2025qwen3}. We set the hyperparameter $\lambda=2$ in Eq.~(\ref{eq:loss}) during training.
Unless otherwise specified, we retrieve 12 semantic and 4 episodic memory entries by default.
All evaluations are conducted on a server equipped with $8\times$ NVIDIA RTX 5880 GPUs. Furthermore, we report {accuracy}, {response latency}, and memory footprint as the evaluation metrics. 
For more experimental results and details, please refer to the Appendix.~\ref{sec:app:exp}.

\begin{figure*}[!h]
  \centering
  \begin{minipage}[b]{0.42\textwidth}
    \centering
    \includegraphics[width=\textwidth]{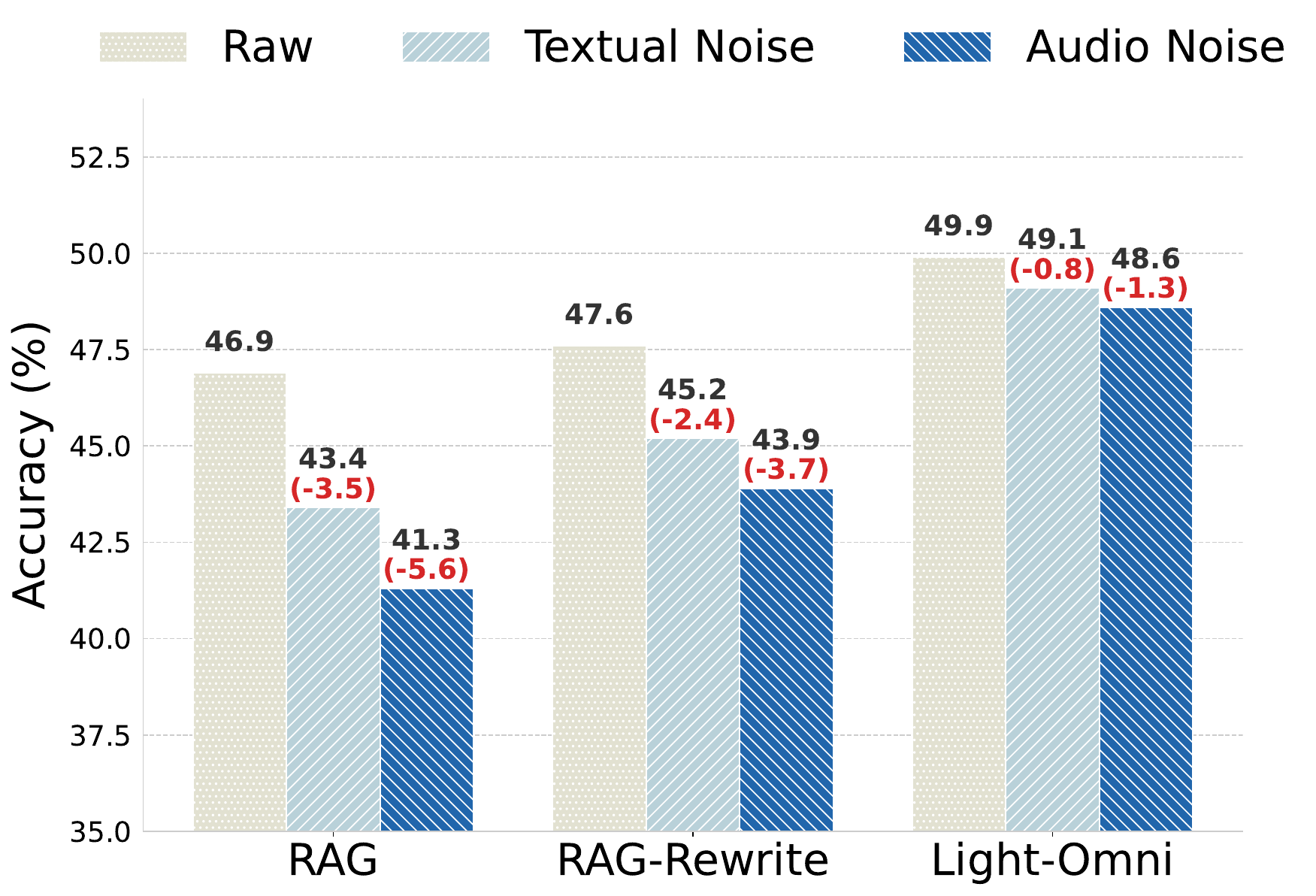}
    \caption{Robustness evaluation under noisy queries on LVBench. Light-Omni shows superior robustness.}
    \label{fig:noise_left}
  \end{minipage}
  \hfill
  \begin{minipage}[b]{0.55\textwidth}
    \centering
    \includegraphics[width=\textwidth]{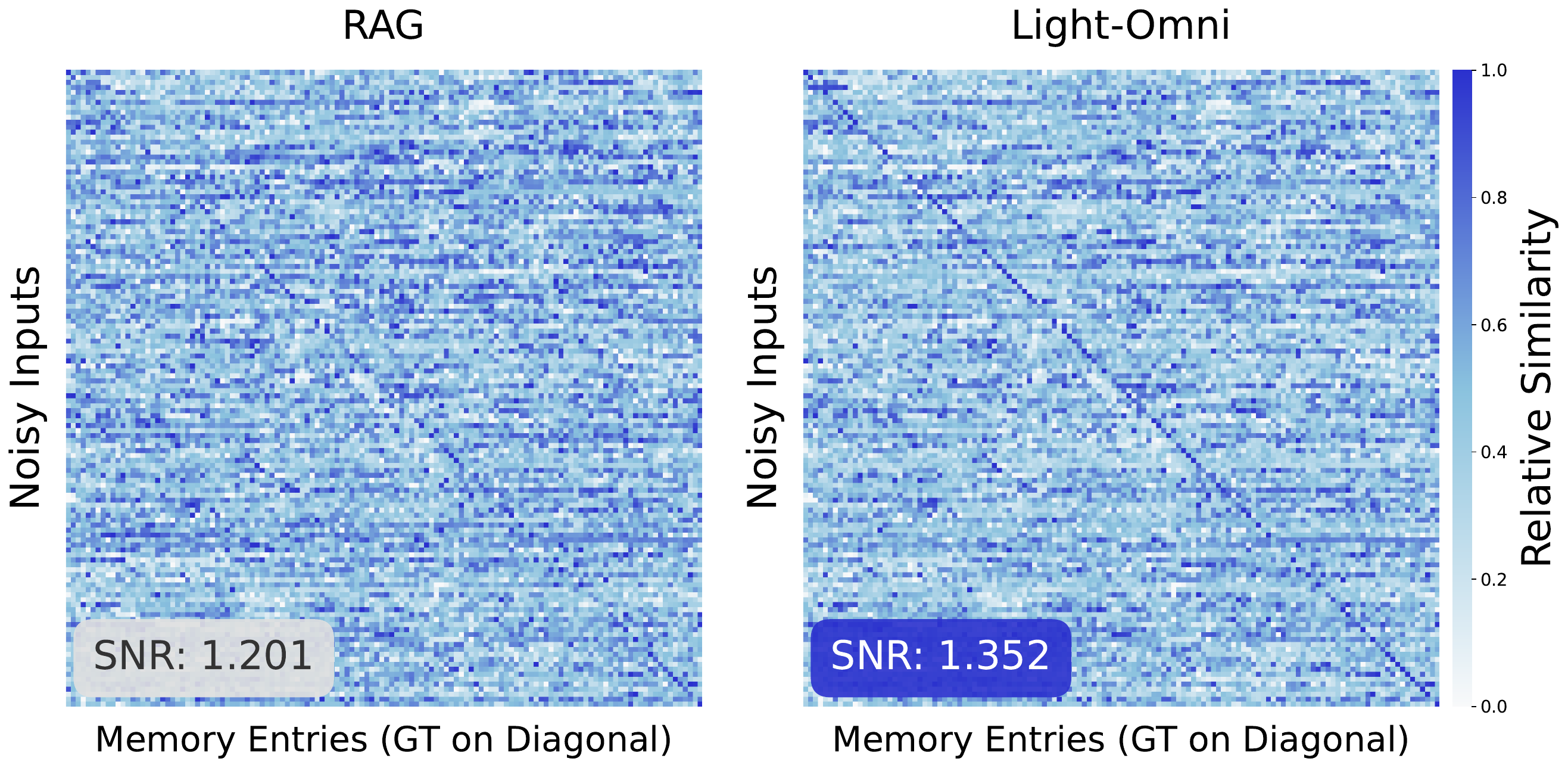}
    \caption{
Similarity matrices between queries and ground-truth memories. Compared to naive RAG, Light-Omni achieves a sharper diagonal and higher SNR, ensuring robust semantic alignment under noisy inputs.}
    \label{fig:noise_right}
  \end{minipage}
\end{figure*}

\begin{table*}[!h]
\centering
\caption{
Efficiency comparison on VideoMME-long. ``Interactive Response'' measures per-query latency, while ``Memorization'' denotes the offline processing cost for an entire video (avg. 2436s). Light-Omni achieves the best accuracy with over 8$\times$ speedup than iterative agents.}
\label{tab:efficiency}
\resizebox{0.99\textwidth}{!}{
\begin{tabular}{ll|cccc|cc|c}
\toprule
\multirow{2}{*}{\textbf{Model}} &\multirow{2}{*}{\textbf{Backbone}}& \multicolumn{4}{c|}{\textbf{Interactive Response}} & \multicolumn{2}{c}{\textbf{Memorization }} & \multirow{2}{*}{\textbf{Acc.}}\\
\cmidrule(lr){3-6} \cmidrule(lr){7-8}
&& Retr. (s) & Resp. (s) & {Total (s)} & Calls & {Total Time (s)} & Calls \\
\midrule
M3-Agent~\cite{long2025seeing} &Qwen3-32B& 16.09 & 9.82 & 25.91 & 2.5 & 3312.78 & 81.2 &61.8 \\
WorldMM~\cite{yeo2025worldmm} &Qwen3-VL-8B& - & - & $\geq$20.00 & $\leq$5.0 & - & - &66.0 \\
\midrule
\rowcolor{blue!20}
\textbf{Light-Omni (ours)}&Qwen2.5-Omni-7B & 0.76 & 1.52 & {2.28} & \textbf{1.0} & {1753.30} & 153.5 &\textbf{66.1}\\
\bottomrule
\end{tabular}
}
\end{table*}

\subsection{Main Results}
\noindent\textbf{Regarding Question 1, Light-Omni exhibits strong performance across all long-video benchmarks.}
As shown in Table~\ref{tab:main_results}, Light-Omni achieves an average accuracy of 58.0\% on VideoMME-long and LVBench, outperforming commercial models like GPT-4o (48.1\%) and Gemini-2.0-Flash (55.8\%), as well as large-scale open-source MLLMs like Qwen2.5-VL-72B (56.1\%).
Compared to the baseline Qwen2.5-Omni-7B, Light-Omni delivers a 9.5\% accuracy gain with a nearly $20.5\times$ speedup and a $3.3\times$ reduction in memory footprint.
In Fig.~\ref{fig:reasoning}, we compare Light-Omni with strong reasoning-based video agents. Light-Omni achieves the best overall performance, surpassing the runner-up WorldMM-8B~\cite{yeo2025worldmm} by 0.9\%. Notably, when expanding the evaluation across the three datasets shown in Fig.~\ref{fig:reasoning} (including HippoVlog), it outperforms M3-Agent~\cite{long2025seeing} by a wider margin of 5.9\%. These results validate the effectiveness and generalizability of Light-Omni.

\vspace{2mm}
\noindent\textbf{To answer Question 2, Light-Omni effectively serves as a multimodal memory system to enhance both the performance and efficiency of existing MLLMs.}
As demonstrated at the bottom of Table~\ref{tab:main_results}, integrating Light-Omni yields consistent accuracy improvements across open-source and commercial models.
Specifically, it boosts the average accuracy of Qwen2.5-VL-7B, Qwen3-VL-8B, and Gemini-2.0-Flash by 4.9\%, 2.5\%, and 3.8\%, respectively.
Beyond performance gains, Light-Omni significantly optimizes resource consumption for models struggling with long contexts.
For instance, when applied to Qwen2.5-VL-7B, it accelerates inference speed by up to 7.2$\times$ and reduces memory footprint by 2.5$\times$ on LVBench. 
These results confirm its strong generalization and viability as a universal memory module.

\vspace{2mm}
\noindent\textbf{In response to Question 3, the dual-state mechanism equips Light-Omni with superior performance in complex scenarios.} 
To evaluate retrieval robustness, we inject noise into the LVBench inputs.
Specifically, we introduce two types of noise that exclusively target the retrieval phase: (1) \textit{Textual Noise}, generated by prepending random text (sourced from the C4 dataset\footnote{https://huggingface.co/datasets/allenai/c4}) four times the length of the original context; and (2) \textit{Audio Noise}, created by synthesizing the text query into speech and integrating it into the final video segment.

As illustrated in Fig.~\ref{fig:noise_left}, Light-Omni exhibits minimal performance degradation compared to standard RAG and RAG-Rewrite. For instance, under audio noise, Light-Omni experiences only a 1.3\% performance degradation, whereas RAG and RAG-Rewrite suffer severe drops of 5.1\% and 3.7\%, respectively.
Furthermore, Fig.~\ref{fig:noise_right} visualizes the similarity matrices between 256 noisy queries and their corresponding ground-truth memory entries. Compared to standard RAG, Light-Omni produces a much sharper and more distinct diagonal, indicating highly precise retrieval of the correct memories despite the noise. The higher Signal-to-Noise Ratio (SNR: 1.352 vs. 1.201) further demonstrates the effectiveness of our dual-state mechanism.

\vspace{2mm}
\noindent\textbf{For Question 4, Light-Omni minimizes system latency and enables dynamic response and retrieval to enhance user experience.}
By decoupling memory consolidation from the active response generation, our framework ensures highly fluid human-agent interactions. 
In Table~\ref{tab:efficiency}, we compare the online latency and offline memory construction overhead of Light-Omni against the multi-turn reasoning baseline, M3-Agent\footnote{The memorization time for M3-Agent excludes API request latency, reporting only local testing time.}.
When generating interactive responses, Light-Omni requires only a single inference call and utilizes an additional prefill stage for retrieval, thereby reducing the overall system latency by {11.36$\times$}.
Furthermore, the memory construction time of M3-Agent exceeds the actual video duration. In contrast, although Light-Omni executes more calls, it incurs significantly lower time overhead, which is attributed to the optimization strategies detailed in Fig.~\ref{fig:latency}.
Beyond efficiency, Light-Omni supports proactive response in real-world scenarios and incorporates user profiles to further enhance personalized experiences. 
Ultimately, this lightweight paradigm eliminates the computational bottlenecks that typically hinder traditional iterative agents, paving the way for seamless, continuous multimodal assistance. Online evaluation on OVO-Bench is detailed in the supplemental~material.

\begin{table}[!t]
\centering
\begin{minipage}[t]{0.5\textwidth}
    \centering
    \captionof{table}{Ablation study of different global state strategies on VideoMME-long~\cite{fu2025video}. ``Vanilla'' denotes that response generation relies solely on retrieved results. ``Uniform'' and ``STM'' indicate using uniformly sampled frames and short-term memory as the global context, respectively.}
    \label{tab:ablation_global_state}
    \setlength{\tabcolsep}{5mm}
    \begin{tabular}{l|cc}
    \toprule
    \textbf{Strategies} & \textbf{Acc. (\%)} &$\triangle$ \\ 
    \midrule
    Vanilla (baseline)           & 63.11 &\\
    \midrule
    Uniform        & 64.11 & {\color{gray}${+1.0}$}\\
    STM            & 63.67 & {\color{gray}${+0.56}$} \\
    \rowcolor{blue!20}
    \textbf{$\mathcal{S}_g$ (ours)} & \textbf{66.10}&{\color{gray}${+2.99}$} \\ 
    \bottomrule
    \end{tabular}
  \end{minipage}
\end{table}
\begin{table}[!h]
\centering
\begin{minipage}[t]{0.5\textwidth}
    \centering
    \captionof{table}{Ablation of retrieval representations on LVBench~\cite{wang2025lvbench}. ``Text Emb.'' and ``Latent Emb.'' denote using the encoded query and latent features for retrieval, respectively. ``Soft Prompt'' refers to jointly encoding the latent features as prefix tokens alongside the query.}
    \label{tab:ablation_latent_state}
    \setlength{\tabcolsep}{5mm}
    \begin{tabular}{l|cc}
    \toprule
    \textbf{Strategies} & \textbf{Acc. (\%)} &$\triangle$ \\ 
    \midrule
    Text Emb. (baseline)     & 49.26 &\\
    \midrule
    Latent Emb.    & 46.48&{\color{gray}${-2.78}$} \\
    Soft Prompt    & 43.45&{\color{gray}${-5.81}$} \\
    \rowcolor{blue!20}
    \textbf{$\mathcal{S}_l$ (ours)} & \textbf{49.90}&{\color{gray}${+0.64}$} \\ 
    \bottomrule
    \end{tabular}
  \end{minipage}
\end{table}

\subsection{Analysis of Dual States}
\noindent\textbf{Global State $\mathcal{S}_g$.}
Functioning as the system's cognitive brain, $\mathcal{S}_g$ empowers Light-Omni to comprehend historical context.
We evaluate it on VideoMME-long~\cite{fu2025video}, which emphasizes global reasoning.
We compare our approach with three common strategies: (1) \textit{Vanilla}, which relies solely on retrieval without context; (2) \textit{Uniform}, which uses uniform sampling; and (3) \textit{STM}, which employs short-term memory as context.
For a fair comparison, we retrieve the same number of topics for all comparative strategies.
As shown in Table~\ref{tab:ablation_global_state}, our hierarchical $\mathcal{S}_g$ achieves the highest accuracy (66.10\%), outperforming \textit{Vanilla} and \textit{Uniform} by 2.99\% and 1.99\%, respectively. This proves that hierarchical memory captures long-term temporal dependencies much more effectively than simplistic sampling or short-term context strategies.

\begin{figure}[!h]
  \centering
  \begin{minipage}[t]{0.5\textwidth}
    \vspace{0pt}
    \centering
    \includegraphics[width=\textwidth]{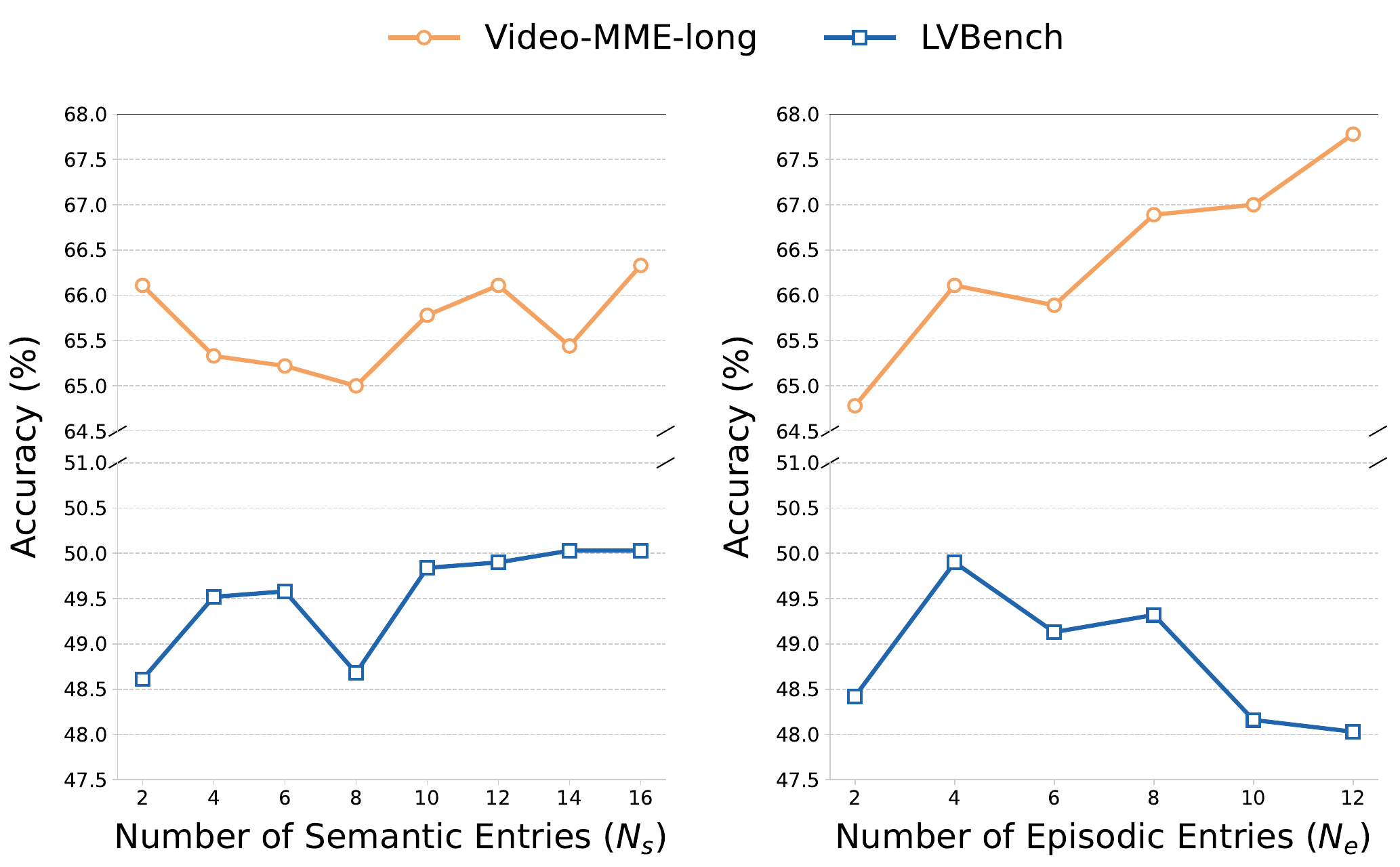}
    \captionof{figure}{Accuracy vs. retrieval granularity. Performance stabilizes around $N_s=12$ for semantic memory. However, for episodic memory, $N_e > 4$ introduces distracting noise on LVBench.}
    \label{fig:retrieval_viz}
  \end{minipage}
\end{figure}

\noindent\textbf{Latent State $\mathcal{S}_l$.}
We evaluate the efficacy of $\mathcal{S}_l$ on the LVBench benchmark~\cite{wang2025lvbench}. 
The comparative strategies include: 
(1) \textit{Text emb.}, which employs standard RAG using only text embeddings; 
(2) \textit{Latent emb.}, which retrieves solely based on latent embeddings; 
(3) \textit{Soft-Prompt}, which prepends latent embeddings to text tokens as soft prompts; and 
(4) \textit{Our $\mathcal{S}_l$}, which fuses text embeddings and latent embeddings via element-wise addition to generate robust representations.
As shown in Table~\ref{tab:ablation_latent_state}, relying exclusively on \textit{Latent emb.} leads to a substantial performance degradation (3.42\%). This is attributed to the difficulty of adapting a generation-pretrained backbone for direct retrieval tasks via lightweight LoRA fine-tuning.
Furthermore, the \textit{Soft-Prompt} strategy yields inferior results (dropping by 5.81\% compared to Text emb.), likely due to the optimization challenges in aligning soft prompts with the frozen embedding space.
In contrast, our proposed $\mathcal{S}_l$ effectively mitigates these limitations, achieving the highest accuracy of 49.90\%. Its robustness is further corroborated under noisy scenarios, as illustrated in Fig.~\ref{fig:noise_left}.

\begin{figure}[!t]
  \centering
  \begin{minipage}[t]{0.5\textwidth}
    \vspace{0pt}
    \centering
    \includegraphics[width=\textwidth]{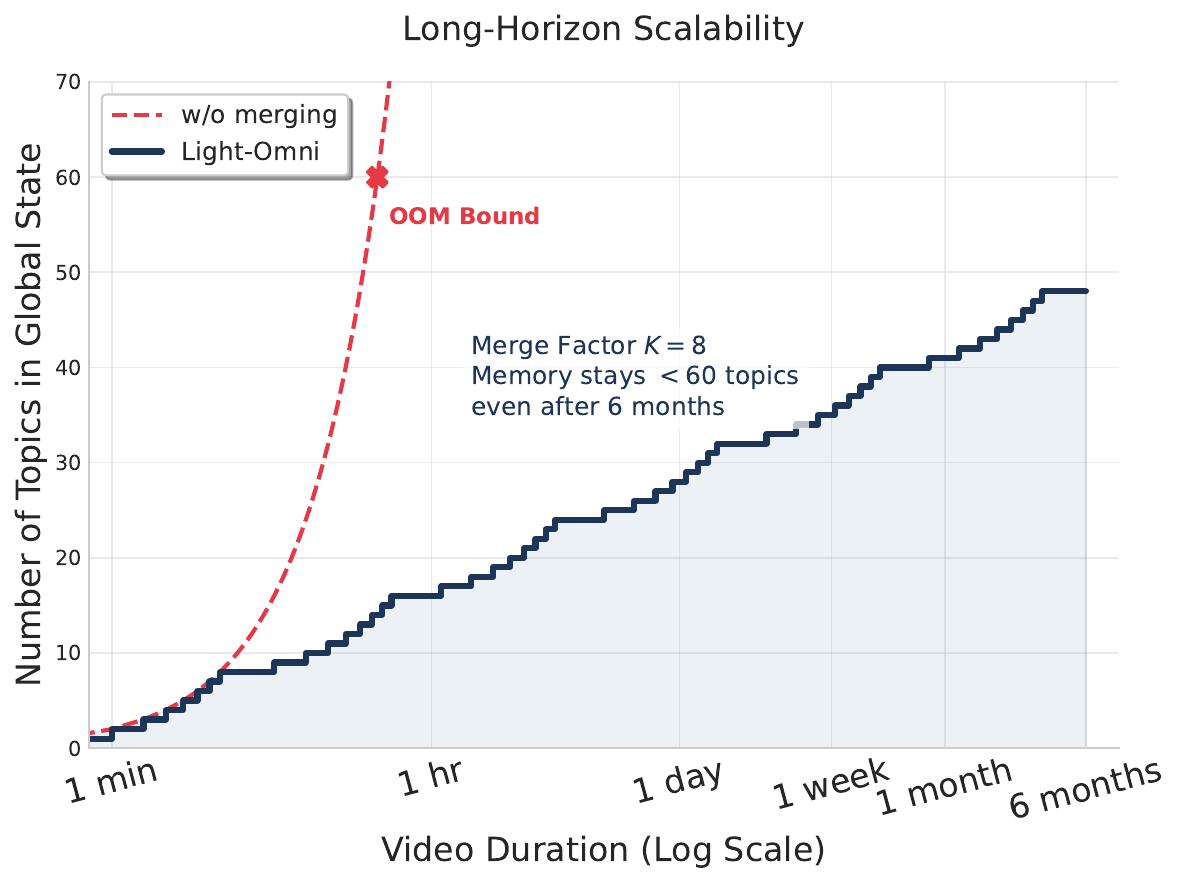}
    \caption{Scalability of Light-Omni over extreme time horizons.}
    \label{app:fig:topic}
  \end{minipage}
  \vspace{-2mm}
\end{figure}

\begin{figure}[!t]
  \centering
  \begin{minipage}[t]{0.5\textwidth}
    \vspace{0pt}
    \centering
    \includegraphics[width=\textwidth]{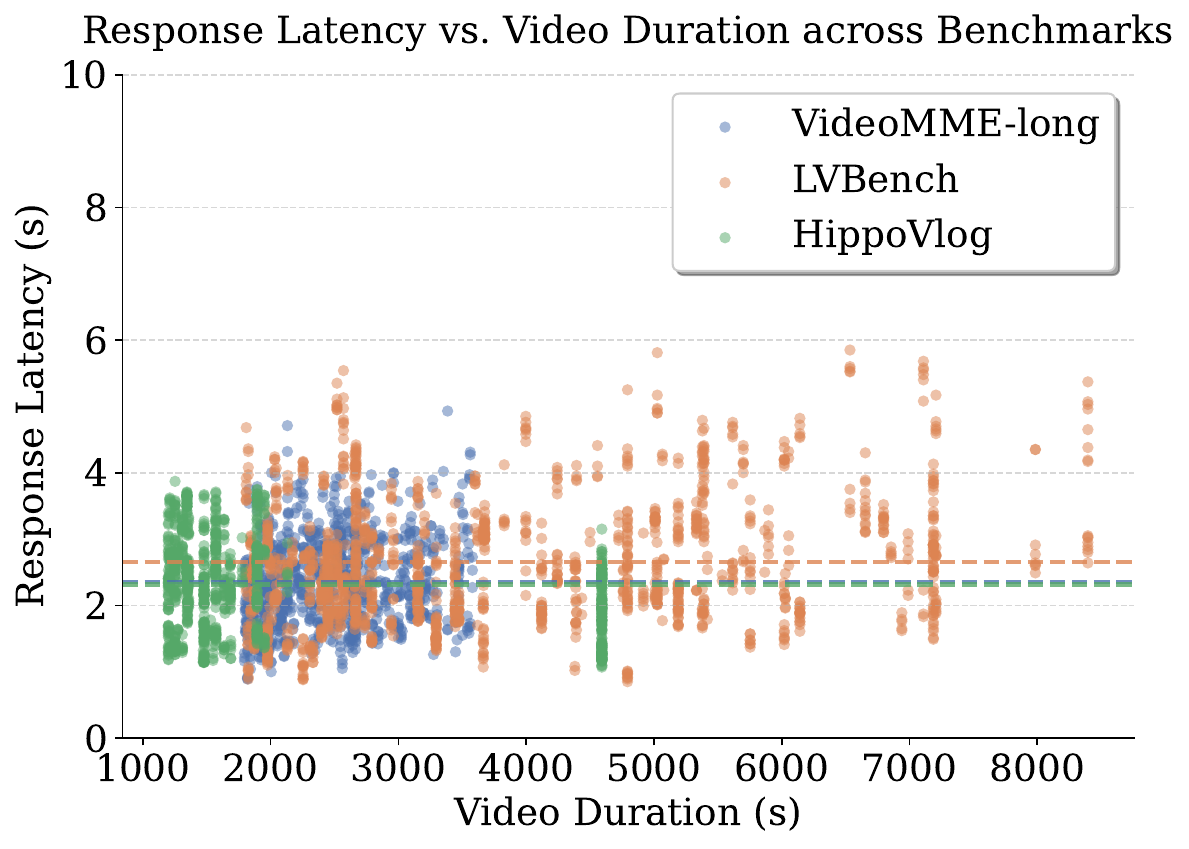}
    \caption{Near-constant response latency of Light-Omni across varying video durations.}
    \label{app:fig:latency}
  \end{minipage}
\end{figure}

\subsection{Influence of Retrieval Granularity}
We analyze how the number of retrieved semantic ($N_s$) and episodic ($N_e$) entries affects performance across VideoMME-long and LVBench (Fig.~\ref{fig:retrieval_viz}).
For semantic memory (left panel), increasing $N_s$ steadily improves accuracy on LVBench and maintains robust performance on VideoMME-long, plateauing around $N_s=12$.
However, a divergent trend emerges for episodic memory: while performance on VideoMME-long continuously benefits from a larger $N_e$, it peaks on LVBench at $N_e=4$ and sharply declines thereafter. This indicates that while semantic facts are generally beneficial, excessive episodic details may act as distracting noise, complicating certain reasoning tasks.
Therefore, we use $N_s=12$ and $N_e=4$ as the default settings.

\subsection{Complexity and Scalability Analysis}
To demonstrate the near-$\mathcal{O}(1)$ inference complexity of Light-Omni, we evaluate its memory and latency scalability across varying temporal horizons. As illustrated in Fig.~\ref{app:fig:topic}, our hierarchical merging strategy ensures a near-constant memory footprint. Notably, even after six months of simulated continuous interaction, the number of topics in the global state remains constrained to merely three times that of a one-hour session, effectively preventing memory~overflow. 

Furthermore, Fig.~\ref{app:fig:latency} demonstrates that Light-Omni maintains a consistently low latency, averaging 2.37 seconds and remaining well below 6 seconds even for two-hour long-form~videos. These results validate that our reflexive dual-state mechanism successfully decouples inference overhead from input duration, thereby providing a highly scalable solution for long-term multimodal~interaction.

\section{Conclusion}
In this work, we present \textbf{Light-Omni}, a novel framework that shifts the paradigm of agentic video understanding from slow, deliberate reasoning to fast, reflexive action.
To bridge the inherent semantic gap between queries and memory representations, we introduce the dual contextual states that enable non-parametric global coherence and semantically aligned evidence retrieval.
Extensive evaluations across multiple long-video benchmarks confirm that Light-Omni outperforms  reasoning-based agents with minimal latency and memory footprint.
Beyond video assistance, Light-Omni's versatile memory framework enhances existing MLLMs, enabling more responsive and practical multimodal agents.

\section*{Acknowledgment}
This work is supported by Fundamental and Interdisciplinary Disciplines Breakthrough Plan of the Ministry of Education of China (JYB2025XDXM902).

%% file: X_suppl.tex
\clearpage
\onecolumn
\begin{center}
    \vspace*{1em}
    \LARGE \textbf{Supplementary Material for} \\
    \vspace{0.5em}
    \Large \textbf{Light-Omni: Reflex over Reasoning in Agentic Video Understanding \\ with Long-Term Memory}
    \vspace{2em}
\end{center}%

\setcounter{section}{0}
\renewcommand{\thesection}{\Alph{section}}
This supplementary material provides additional details to complement the main paper, structured as follows:

\begin{itemize}
\item Appendix~\ref{sec:app:memory} elaborates on the proposed multimodal memory system, including its storage and retrieval mechanisms, as well as the current memory capabilities.

\item Appendix~\ref{sec:app:data} details the training data, including its construction pipeline and statistics.

\item Appendix~\ref{sec:app:opt} presents comprehensive training details and the online inference pipeline.

\item Appendix~\ref{sec:app:exp} provides additional experiments and analyses to show the superiority of Light-Omni.

\item Appendix~\ref{sec:app:dis} discusses the potential impacts and limitations of Light-Omni.

\item Appendix~\ref{sec:app:prompt} outlines the system prompts utilized by Light-Omni across different stages.
\end{itemize}

\section{Multimodal Memory System}
\label{sec:app:memory}
As described in Section~\ref{sec:method:memory}, the proposed multimodal memory system of Light-Omni consists of three distinct components. This section provides a comprehensive elaboration of their storage formats and update mechanisms.

\subsection{Memory Storage and Management}

\noindent\textbf{User Profile ($\mathcal{M}_p$):} This component stores identity-related information extracted from the input multimodal stream, encompassing user avatars, preferences, and behavioral traits, which are crucial for delivering personalized experiences in Light-Omni.
Specifically, for the incoming visual stream (sampled at 1 fps), we employ \texttt{Buffalo\_s}\footnote{https://huggingface.co/immich-app/buffalo\_s} for face detection and recognition.
To maintain identity consistency and reduce false detections across frames, we implement a feature-level tracking mechanism coupled with an exponential moving average for dynamic facial embedding updates.
Subsequently, we utilize constructed memory topics to update the profile entities involved in $k=8$ consecutive segments.
This updating phase occurs concurrently with the merging of the $\mathcal{M}_e$.

\vspace{2mm}
\noindent\textbf{Semantic Memory ($\mathcal{M}_s$):} This component distills factual knowledge, relationships, and abstract concepts from each input segment.
$\mathcal{M}_s$ operates as an append-only structure, continuously adding new memory entries derived from the generated $\text{Topic}^t$ at time step $t$. Each entry is formatted as ``\texttt{[start\_timestamp]--[end\_timestamp]: content}''. To ensure memory density and avoid redundancy, a content-based filtering mechanism is employed prior to appending new entries.

\vspace{2mm}
\noindent\textbf{Episodic Memory ($\mathcal{M}_e$):} This component archives all past events and raw multimodal inputs in strictly chronological order.
Notably, it adopts a narrative-style incremental addition approach, which preserves temporal continuity while preventing the computational overhead caused by storing highly repetitive visual frames.
Similar to $\mathcal{M}_s$, episodic entries are extracted from $\text{Topic}^t$ and encompass comprehensive contextual logs: start time, end time, visual descriptions, auditory descriptions, assistant responses (if applicable), and pointers to the raw multimodal inputs.

As illustrated in Table~\ref{app:tab:memory_examples}, we provide concrete examples for these three types of memory.
To establish robust cross-references and semantic connections between different memory modules, we utilize unified specific identifiers, such as \texttt{<face\_idx>}, to represent recognized individuals globally.

\begin{table*}[htbp]
\centering
\footnotesize 
\renewcommand{\arraystretch}{1.4}
\caption{Examples of the three components in the Multimodal Memory System. We use \texttt{<face\_idx>} to establish cross-references between entities and maintain identity consistency.}
\label{app:tab:memory_examples}
\begin{tabular}{>{\raggedright\arraybackslash}p{0.25\linewidth} | >{\raggedright\arraybackslash}p{0.27\linewidth} | >{\raggedright\arraybackslash}p{0.4\linewidth}}
\toprule
\textbf{User Profile ($\mathcal{M}_p$)} & \textbf{Semantic Memory ($\mathcal{M}_s$)} & \textbf{Episodic Memory ($\mathcal{M}_e$)} \\
\midrule
\textbf{Entity:} \texttt{<face\_0>} \newline
\textbf{Name:} Dr. Jason Fung \newline
\textbf{Demographics:} Adult male, Middle-aged, Wears glasses. \newline
\textbf{Occupation:} Medical doctor, researcher, educator. \newline
\textbf{Persona:} Critical of conventional medicine, advocates for intermittent dietary management and hormonal obesity theory.
& 
\texttt{2026-01-01 08:28:11} \newline
\texttt{-- 2026-01-01 08:28:40:} \newline 
Mr. WK has been on insulin since 2007. 
\vspace{1em} \newline
\texttt{2026-01-01 08:00:00} \newline
\texttt{-- 2026-01-01 08:00:29:} \newline 
\texttt{<face\_0>} states that diabetes medications lose effectiveness over the years.
& 
\texttt{2026-01-01 08:00:00--08:00:29} \newline
\textbf{Visual:} Scene opens in a bright, modern medical office with a large window. \texttt{<face\_0>} sits at a desk wearing a white lab coat over a blue shirt, gesturing with his hands. A screen displays a medical slide titled `THE ANCIENT ...'. \newline
\textbf{Audio:} \texttt{<face\_0>} (informative): ``What happens is that the diabetes they tell you is a chronic disease... As you take medications...'' \newline
\textbf{Assistant:} \textit{null} \\
\bottomrule
\end{tabular}
\end{table*}

\subsection{Memory Retrieval and Utilization}

During the Light-Omni runtime, we dynamically retrieve the required memory components based on the current context.
Specifically, for the User Profile ($\mathcal{M}_p$), the system retrieves only the profiles associated with the individuals (faces) actively detected in the current visual input.
Conversely, for Semantic ($\mathcal{M}_s$) and Episodic Memory ($\mathcal{M}_e$), the retrieval pipeline is conditionally activated only when the \texttt{speech} action is triggered. 
We employ the Qwen3-Embedding-0.6B~\cite{zhang2025qwen3} to encode the memory entries into dense embeddings.
In voice-based interaction scenarios, we utilize Whisper\footnote{https://huggingface.co/openai/whisper-small} to transcribe the user's audio stream into text.
The dense embedding of this textual transcription is then fused with the current latent embeddings to formulate the query embedding.
Finally, to ensure low-latency responsiveness during online inference, the entire dense retrieval process is accelerated using Facebook AI Similarity Search (FAISS)~\cite{johnson2019billion}.

\subsection{Memory Capabilities Supported by Light-Omni}
The current version of Light-Omni features five primary memory-driven capabilities.
To process continuous input streams efficiently, we employ the \texttt{silero-vad}\cite{SileroVAD} toolkit for Voice Activity Detection (VAD).
By restricting the maximum segment length to 30 seconds, this design ensures low-latency processing and empowers the agent with proactive interaction capabilities over long horizons.

\begin{itemize}
\item \textbf{Historical Visual and Auditory Retrieval:} The system retains comprehensive logs of past multimodal states, encompassing visual occurrences (e.g., environments, human actions) and auditory details.

\item \textbf{Personalized Identity, Relationships, and Preferences:} The system maintains user-centric profiles powered by $\mathcal{M}_p$, mapping recognized individuals to their specific demographic traits, social relationships, habits, and personalized preferences.

\item \textbf{Episodic Event Summarization:} The system abstracts and summarizes long-term historical events, transforming raw multimodal streams into high-level records.

\item \textbf{Active and Passive Memorization:} The system dynamically captures knowledge through both explicit user instructions (e.g., actively instructed to ``Remember that...'') and implicit conversational context, ensuring a comprehensive understanding of user intents.

\item \textbf{Proactive Scheduling and Output:} The system handles time-aware tasks proactively, supported by the chronological nature of $\mathcal{M}_e$. For instance, if a user requests, ``Remind me to [Task] in three minutes,'' the model can track the time, trigger the event, and proactively deliver the response at the designated timestamp.
\end{itemize}

\begin{figure*}[!t]
    \centering
    \includegraphics[width=\linewidth]{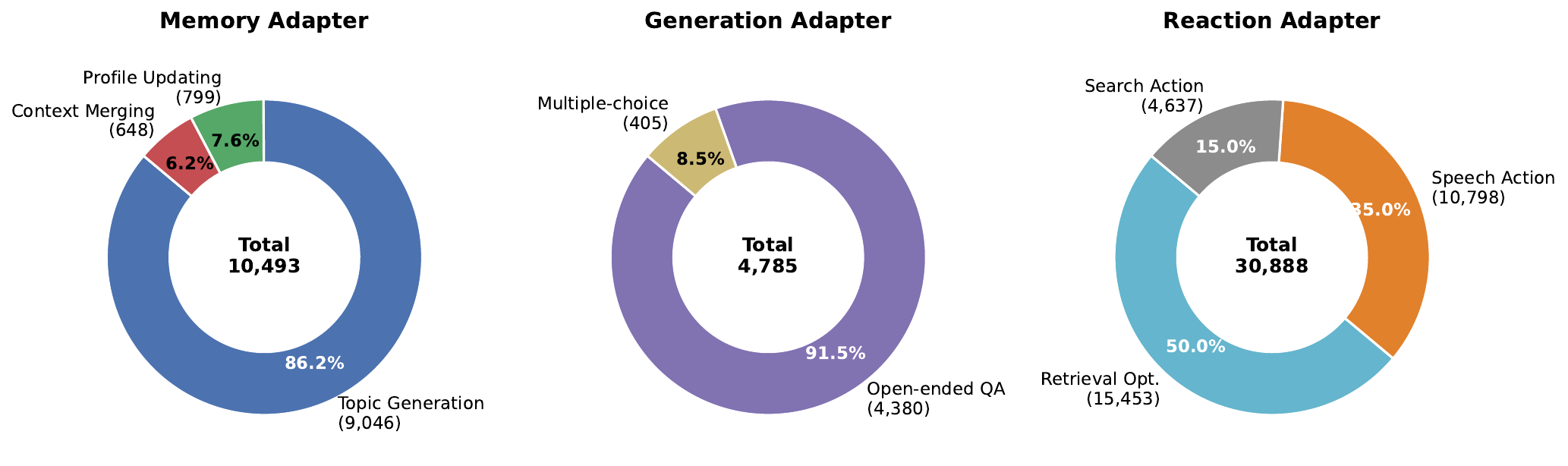}
    \caption{Distribution of the Light-Omni training dataset across the three adapters. The dataset is heavily weighted towards the Memory and Reaction adapters to cultivate agentic behaviors.}
    \label{fig:data_dist}
\end{figure*}

\section{Training Dataset Construction}
\label{sec:app:data}
Existing open-source video datasets primarily focus on offline understanding and inherently lack continuous, long-horizon interactive contexts~\cite{long2025seeing}. To bridge this gap, we design a novel data synthesis pipeline to construct a tailored training set for Light-Omni. The construction proceeds in four sequential steps: 
(i) \textbf{Video Collection \& Grouping:} We initially collect public videos from platforms like YouTube and randomly group them to simulate prolonged, multi-session user interactions. 
(ii) \textbf{User Query Insertion:} Simulated user dialogues are injected at various temporal anchor points within the video streams. We utilize CosyVoice~\cite{du2024cosyvoice} for text-to-speech synthesis, seamlessly blending the generated audio into the original soundtrack while meticulously preserving acoustic tones and background noise to maximize environmental realism. 
(iii) \textbf{Response Generation:} To generate ground-truth assistant responses, we prompt a commercial Large Multimodal Model (specifically, \texttt{Gemini-3-Flash-Preview}, chosen to balance performance and cost) to act as a personalized assistant provided with the complete context. 
(iv) \textbf{Intermediate Signal Construction:} Finally, we extract intermediate signals encompassing the memory construction and retrieval processes, perfectly aligning with the pipeline depicted in Fig.~\ref{fig:fra}.

As illustrated in Fig.~\ref{fig:data_dist}, this comprehensive synthesis pipeline yields a total of 46,166 training samples derived from 2,118 unique videos. The dataset is strategically partitioned into three subsets to train distinct capabilities:
\begin{itemize}
    \item \textbf{Memory Adapter (10,493 samples):} Comprising 9,046 for topic generation, 648 for context merging, and 799 for profile updating.
    \item \textbf{Generation Adapter (4,785 samples):} Dedicated to fundamental QA, including 4,380 for open-ended QA and 405 for multiple-choice tasks.
    \item \textbf{Reaction Adapter (30,888 samples):} Tailored for latent state generation, encompassing 10,798 for speech action, 4,637 for search action, and 15,453 for retrieval optimization.
\end{itemize}
By design, we allocate a substantially larger proportion of data to the memory and reaction adapters to effectively imbue the backbone model with novel agentic capabilities. Conversely, the volume of generation samples is kept relatively minimal to preserve the model's inherent general-purpose capabilities without catastrophic forgetting.

\section{Optimization Details and Runtime Pipeline}
\label{sec:app:opt}

\subsection{Training Details}
We implement the optimization pipeline of Light-Omni utilizing the open-source \texttt{ms-swift}\footnote{https://github.com/modelscope/ms-swift} framework.
To effectively adapt the Qwen2.5-Omni-7B backbone to various system capabilities, the different adapters are optimized independently. 
During the training process, the adapters are optimized using Low-Rank Adaptation (LoRA)~\cite{hu2022lora} with a rank of $r=64$, $\alpha=128$, and a dropout rate of $0.05$.
These adapters comprehensively target all linear projection modules.
The entire optimization procedure is conducted on a single computing node equipped with 8 NVIDIA H800 GPUs, and the total training time is approximately 10 hours. The hyperparameter configurations for the independent adapter training are detailed in Table~\ref{app:tab:hyperparams}.

\begin{algorithm}[!t]
\caption{Runtime Pipeline of Light-Omni}
\label{app:alg}
\begin{algorithmic}[1]
\renewcommand{\algorithmicrequire}{\textbf{Input:}}
\Require Continuous stream $\mathcal{I}$, backbone $\pi_\theta$, memory system $\mathcal{M}$, queue $Q$.

\Statex \textcolor{gray}{// Real-Time Reflex}
\While{Stream $\mathcal{I}$ is active}
    \State $\mathcal{I}^t \leftarrow$ Execute data \textit{preprocessing} on current segment
    \If{end-of-speech detected \textbf{or} interval $> 30s$}
        \State $\mathbf{H}^t \leftarrow \pi_\theta(\mathcal{S}_g^t, \mathcal{I}^t, \mathbf{P}_{soft})$ \Comment{Forward pass}
        \State $\mathcal{S}_l^t=\{h_{act}^t, h_{ret}^t\}$ \Comment{Obtain latent state}
        \State $a_{act}^t, z_{ret}^t \leftarrow \text{Decode}(\mathcal{S}_l^t)$ via Eq.~(\ref{eq:latent_generation}) \Comment{Determine action and retrieval embedding}
        \State $\mathcal{R}^t \leftarrow \text{Execute}(a_{act}^t, \text{Retrieve}(\mathcal{M}^t, z_{ret}^t))$ \Comment{Generate response}
        \State \textbf{Push} $(\mathcal{I}^t, \mathcal{R}^t)$ \textbf{into} Queue $Q$ \Comment{Enqueue for lazy update}
    \EndIf
\EndWhile
\Statex \textcolor{gray}{// Asynchronous Memory Consolidation}
\While{Queue $Q$ is \textbf{not} empty}
    \State Pop $(\mathcal{I}^t, \mathcal{R}^t)$ from $Q$
    \State Generate $\text{Topic}^t = \{\mathcal{M}_s^t, \mathcal{M}_e^t\}$ based on the interaction
    \State Selectively update User Profile $\mathcal{M}_p$ and Global State $\mathcal{S}_g$
\EndWhile
\end{algorithmic}
\end{algorithm}

\begin{table*}[!t]
\centering
\small
\renewcommand{\arraystretch}{1.2}
\caption{Hyperparameter configurations for the independent training of the three Light-Omni adapters.}
\label{app:tab:hyperparams}
\begin{tabular}{l | c | c | c}
\toprule
\textbf{Configuration} & \textbf{Memory Adapter} & \textbf{Generation Adapter} & \textbf{Reaction Adapter} \\
\midrule
LoRA Configuration & \multicolumn{3}{c}{$r=64$, $\alpha=128$, dropout=0.05, target=all-linear} \\
Learning Rate & \multicolumn{3}{c}{$2 \times 10^{-5}$} \\
LR Scheduler & \multicolumn{3}{c}{Cosine (min\_lr\_rate=0.1, warmup\_ratio=0.05)} \\
\midrule
Training Epochs & 2 & 2 & 3 \\
Global Batch Size & 8 & 8 & 16 \\
Training Steps & $2,622$ & $1,196$ & $5,793$ \\
\bottomrule
\end{tabular}
\end{table*}

\subsection{Runtime Pipeline}
For experimental evaluation, Light-Omni follows the workflow depicted in Fig.~\ref{fig:fra}, where memory updates occur immediately after the response generation. 
For online deployment, to prevent memory consolidation operations from impacting response latency, we implement an asynchronous architecture comprising three processes:
(1) The online service continuously performs input preprocessing, including speech activity detection and face recognition;
(2) Upon detecting the end of speech or when the silence interval exceeds 30 seconds, Light-Omni generates reflexive actions and responses, then pushes the interaction data into a queue;
(3) A background process monitors this queue to asynchronously update the memory system.
The detailed operational workflow is outlined in Algorithm~\ref{app:alg}.

\section{Additional Experimental Results and Analysis}
\label{sec:app:exp}

\subsection{Online Video Understanding}
We evaluate Light-Omni's online performance on the Real-Time Visual Perception and Backward Tracing tasks of OVO-Bench~\cite{niu2025ovo}. As shown in Table~\ref{app:tab:ovo_results}, Light-Omni achieves a 54.51\% average accuracy, significantly outperforming open-source online MLLMs such as Dispider~\cite{qian2025dispider} (+3.05\%) and Flash-VStream~\cite{zhang2024flash} (+25.41\%). Furthermore, we provide an interactive demo on our project website, showcasing Light-Omni's robust capabilities in real-time video interaction and continuous multimodal memory in practical scenarios.

\begin{table*}[t]
\centering
\caption{Detailed evaluation results on OVO-Bench~\cite{niu2025ovo}. Accuracy (\%) is reported for each task. The best results among open-source models are highlighted in \textbf{bold}.}
\label{app:tab:ovo_results}
\resizebox{\textwidth}{!}{
\begin{tabular}{l|c|cccccc|c|ccc|c|c}
\toprule
\multirow{2}{*}{\textbf{Model}} & \multirow{2}{*}{\textbf{\# Frames}} & \multicolumn{7}{c|}{\textbf{Real-Time Visual Perception}} & \multicolumn{4}{c|}{\textbf{Backward Tracing}} & \multirow{2}{*}{\textbf{Avg.}} \\
\cmidrule(lr){3-9} \cmidrule(lr){10-13}
 & & OCR & ACR & ATR & STU & FPD & OJR & \textbf{Avg.} & EPM & ASI & HLD & \textbf{Avg.} & \\
\midrule
VideoLLM-online~\cite{chen2024videollm} & 2fps & 8.05 & 23.85 & 12.07 & 14.04 & 45.54 & 21.20 & 20.79 & 22.22 & 18.80 & 12.18 & 17.73 & 20.28 \\
Flash-VStream~\cite{zhang2024flash} & 1fps & 25.50 & 32.11 & 29.31 & 33.71 & 29.70 & 28.80 & 29.86 & 36.36 & 33.78 & 5.91 & 25.35 & 29.10 \\
Dispider~\cite{qian2025dispider} & 1fps & 57.72 & 49.54 & \textbf{62.07} & \textbf{44.94} & 61.39 & 51.63 & 54.55 & \textbf{48.48} & 55.41 & 4.30 & 36.06 & 51.46 \\
\rowcolor{blue!20}
\textbf{Light-Omni (ours)} & 1fps & \textbf{65.10} & \textbf{63.30} & 51.72 & 37.50 & \textbf{70.30} & \textbf{60.33} & \textbf{56.77} & 43.95 & \textbf{67.86} & \textbf{43.24} & \textbf{50.00} & \textbf{54.51} \\
\bottomrule
\end{tabular}
}
\end{table*}



\subsection{Comparison with advanced Video Agents}
In Table~\ref{app:tab:main_results}, we provide a more comprehensive performance comparison between Light-Omni and various leading baselines, including proprietary MLLMs, RAG-based agents, and reasoning-based agentic frameworks.
The comparison data is sourced from~\cite{yeo2025worldmm}.
Light-Omni significantly outperforms all existing agentic baselines, achieving a substantial average accuracy gain of $+10.9\%$ over the baseline (Qwen2.5-Omni-7B~\cite{xu2025qwen2}).
Light-Omni consistently surpasses sophisticated reasoning-based agents such as M3-Agent~\cite{long2025seeing} and {WorldMM-8B}~\cite{yeo2025worldmm}, particularly excelling on {HippoVlog} with 13.0\% and 8.8\% improvements.
These results underscore the effectiveness of our method, offering a superior balance of high accuracy and low computational overhead compared to both retrieval-augmented and iterative-reasoning paradigms.

\begin{table*}[!t]
\centering
\small
\renewcommand{\arraystretch}{1.0}
\caption{Performance comparison on three long video benchmarks.}
\label{app:tab:main_results}
\begin{tabular}{llcccc}
\toprule
\textbf{Model} & & \textbf{Video-MME-long} & \textbf{LVBench} & \textbf{HippoVlog} & \textbf{Avg. Acc.} \\
\midrule
\multicolumn{6}{l}{\textit{\textbf{Commercial \& Open-Source MLLMs}}} \\ \addlinespace[0.1cm]
GPT-5~\cite{singh2025openai} &-& \textbf{74.3} & \textbf{60.4} & 75.7 & \textbf{70.1} \\
Gemini 2.5 Pro~\cite{comanici2025gemini}&- & 55.7 & 57.0 & 72.0 & 61.6 \\
Qwen3-VL-8B~\cite{bai2025qwen3} &8B& 61.0 & 48.3 & 74.4 & 61.2 \\
Qwen2.5-Omni-7B~\cite{xu2025qwen2} &7B& 55.3 & 41.6 & 62.0 & 53.0 \\
\midrule
\multicolumn{6}{l}{\textit{\textbf{RAG-based Video Agents}}} \\ \addlinespace[0.1cm]
LightRAG~\cite{guo2024lightrag} &-& 46.6 & 30.4 & 47.4 & 41.5 \\
Video-RAG~\cite{luo2024video} &-& 55.4 & 33.1 & 65.1 & 51.2 \\
\midrule
\multicolumn{6}{l}{\textit{\textbf{Reasoning-based Video Agents}}} \\ \addlinespace[0.1cm]
Ego-R1~\cite{tian2025ego} &3B& 42.7 & 34.1 & 58.8 & 45.2 \\
HippoMM~\cite{lin2025hippomm} &-& 41.6 & 38.2 & 71.9 & 50.6 \\
M3-Agent~\cite{long2025seeing} &32B& 61.8 & 49.3 & 65.5 & 58.9 \\
WorldMM-8B~\cite{yeo2025worldmm} &8B& 66.0 & \textbf{55.4} & 69.7 & 63.7 \\
\midrule
\rowcolor{blue!20} 
\textbf{Light-Omni (ours)} &7B& \textbf{66.1}{\color{gray}$_{+10.8}$} & 49.9{\color{gray}$_{+8.3}$} & \textbf{78.5}{\color{gray}$_{+16.5}$} & \textbf{64.8}{\color{gray}$_{+11.8}$} \\
\bottomrule
\end{tabular}
\end{table*}

\subsection{Ablation Analysis}
We conduct a comprehensive ablation study to evaluate the contributions of Light-Omni's core components (Table~\ref{app:tab:ablation}).
The results indicate that the global state primarily impacts inference latency, reducing it from 2.37s to 0.81s when removed. 
Moreover, episodic memory is the key driver of overall performance, as its absence causes the largest accuracy drop (from 64.84\% to 60.50\%).
Furthermore, although the latent state provides a modest average gain of 0.68\%, its critical role in retrieval robustness under noisy inputs is further substantiated in Fig.~\ref{fig:noise_left}.

\begin{table*}[!t]
\centering
\small
\renewcommand{\arraystretch}{1.2}
\caption{Ablation study of Light-Omni components. We report Accuracy (\%) and Latency (s) across three long-video benchmarks.}
\label{app:tab:ablation}
\setlength{\tabcolsep}{3.5pt}
\begin{tabular}{l | cc | cc | cc | cc}
\toprule
\multirow{2}{*}{\textbf{Configuration}} & \multicolumn{2}{c|}{\textbf{VideoMME-long}} & \multicolumn{2}{c|}{\textbf{LVBench}} & \multicolumn{2}{c|}{\textbf{HippoVlog}} & \multicolumn{2}{c}{\textbf{Overall Avg.}} \\
\cmidrule(lr){2-3} \cmidrule(lr){4-5} \cmidrule(lr){6-7} \cmidrule(lr){8-9}
& Acc. ($\uparrow$) & Lat. ($\downarrow$) & Acc. ($\uparrow$) & Lat. ($\downarrow$) & Acc. ($\uparrow$) & Lat. ($\downarrow$) & Acc. ($\uparrow$) & Lat. ($\downarrow$) \\
\midrule
w/o Global State      & 63.11 & \textbf{0.83} & 48.87 & \textbf{0.81} & 76.70 & \textbf{0.80} & 62.89 & \textbf{0.81} \\
w/o Latent State      & 65.22 & 2.22 & 49.26 & 2.59 & 78.00 & 2.32 & 64.16 & 2.38 \\
w/o Semantic Memory   & 64.44 & 2.06 & 48.55 & 2.42 & 77.90 & 2.17 & 63.63 & 2.21 \\
w/o Episodic Memory   & 64.11 & 1.62 & 45.38 & 2.04 & 72.00 & 1.76 & 60.50 & 1.81 \\
\midrule
\rowcolor{blue!20} 
\textbf{Light-Omni (ours)} & \textbf{66.11} & 2.22 & \textbf{49.90} & 2.58 & \textbf{78.50} & 2.32 & \textbf{64.84} & 2.37 \\
\bottomrule
\end{tabular}
\end{table*}

\subsection{Redundancy Pruning Analysis}
As detailed in Section~\ref{sec:method:opt}, we employ a redundancy pruning technique to process input visual streams.
This strategy operates at two granularities to mitigate both temporal and spatial redundancy:
(i) \textbf{Inter-frame pruning}, which evaluates temporal similarity by calculating the cosine distance between the mean feature embeddings of consecutive frame tokens; and
(ii) \textbf{Intra-frame pruning}, which applies saliency-based ranking to identify and retain the most informative spatial tokens within each frame.
Specifically, frames identified as temporally redundant are pruned into a minimal anchor set of 32 tokens to preserve essential context, while salient tokens in non-redundant frames are adaptively sampled based on the compression hyperparameter $r_{comp}=0.25$.

\begin{figure*}[t]
\centering
\includegraphics[width=.99\textwidth]{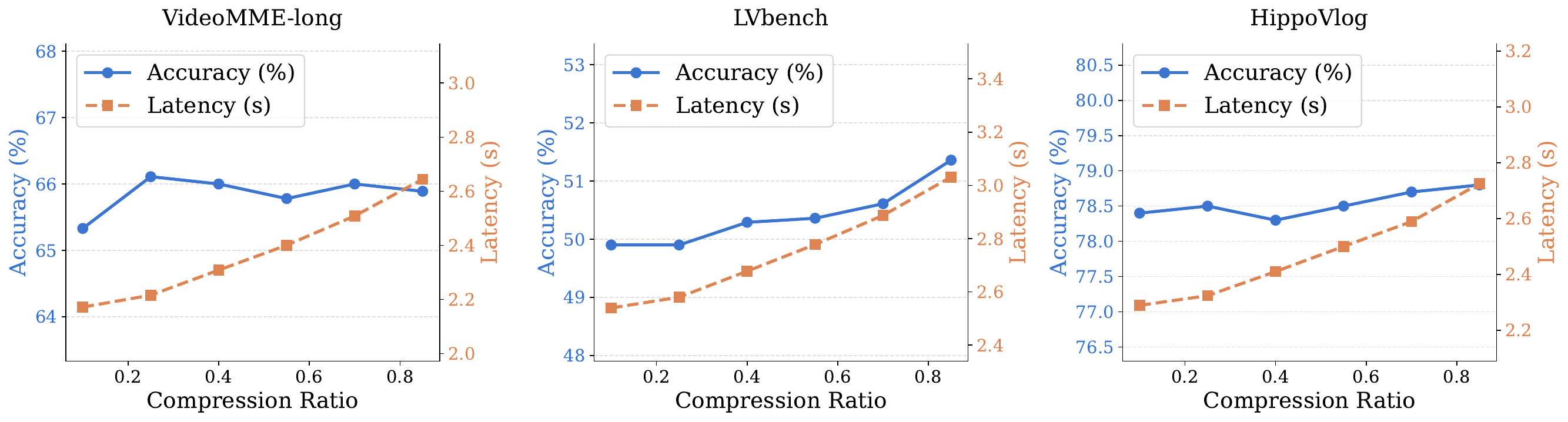} 
\caption{Empirical analysis of performance and system latency across different compression ratios.}
\label{app:fig:ratio}
\end{figure*}

Figure~\ref{app:fig:ratio} illustrates the trade-off between the token retention ratio and system performance across three benchmarks.
We observe that while increasing the ratio generally yields modest accuracy improvements, it incurs a linearly increasing response latency.
Notably, on VideoMME-long, accuracy remains highly resilient even at lower compression ratios.
This confirms that our pruning strategy effectively eliminates over 75\% of redundant visual tokens without compromising the model's core understanding capabilities.

\section{Further Discussion}
\label{sec:app:dis}

\noindent\textbf{Authenticity of the Synthesized Dataset.} To ensure the synthesized dataset faithfully reflects real-world interactive scenarios, we prioritize two design principles.
First, the source videos span diverse domains and temporal contexts.
Second, the simulated in-situ user dialogues undergo meticulous crafting; the synthesized audio mimics the acoustic tones and environmental noise of the original videos to maximize realism. The robust empirical results across multiple long-video benchmarks corroborate the validity of this synthetic pipeline and our overall framework.

\vspace{2mm}
\noindent\textbf{Broader Impact.} This work posits that the cornerstone of video understanding is effective context maintenance and retrieval, rather than heavy reasoning designed to bridge semantic gaps between noisy inputs and memory representations.
Crucially, precise retrieval hinges on non-parametric global context construction and semantically aligned evidence aggregation, which Light-Omni achieves through its dual contextual states.
Our experiments consistently substantiate the superiority of reflexive intelligence over iterative reasoning paradigms. Looking forward, we envision a paradigm shift where foundation models serve as unified architectures for memory representation and retrieval, jointly optimized on long-horizon datasets and tasks, paving the way for more efficient and scalable video understanding systems.

\vspace{2mm}
\noindent\textbf{Limitations.} Despite its advantages, Light-Omni has several limitations.
First, its performance is inevitably constrained by the backbone model; currently, Qwen2.5-Omni-7B lags behind the newer Qwen3 series in certain general capabilities~\cite{Qwen3Omni}.
Nonetheless, as demonstrated in Table~\ref{tab:main_results}, utilizing Light-Omni purely as a plug-and-play memory framework still boosts the average performance of Qwen3-VL-8B by $2.5\%$ across VideoMME-long and LVBench.
Second, the offline memory construction remains computationally expensive.
Although Light-Omni reduces the offline overhead by half compared to M3-Agent~\cite{long2025seeing}, processing massive video streams still demands substantial computational resources.

\section{System Prompts for Light-Omni Framework}
\label{sec:app:prompt}
To facilitate reproducibility regarding our experimental setup, we show the system prompts utilized across various stages of the Light-Omni framework. Specifically, Figs.~\ref{fig:prompt_mem}, \ref{fig:prompt_gen}, \ref{fig:prompt_profile}, and \ref{fig:prompt_global} present the prompts employed for memory construction, response generation, user profile updating, and global state consolidation, respectively.

\definecolor{myThemeDarkBlue}{RGB}{70, 80, 220}
\newtcolorbox{myfigbox}[1][]{
    colframe=myThemeDarkBlue,
    colback=myThemeDarkBlue!5!white,
    fonttitle=\bfseries\small,
    top=1pt,
    bottom=1pt,
    left=2pt,
    right=2pt,
    boxsep=2pt,
    boxrule=1pt,
    arc=3pt,
    outer arc=2pt,
    enhanced,
    #1
}
\lstset{
    numbers=none,
    language={},
    basicstyle=\small\ttfamily,
    aboveskip=0pt,
    belowskip=0pt,
    breaklines=true,
    columns=fullflexible,
    keepspaces=true,
    showstringspaces=false,
    frame=none,
    breakautoindent=false, 
    breakatwhitespace=true,
    breakindent=0pt,
    moredelim=[s][\bfseries]{\{}{\}},
    literate=
        {_}{\_}{1}
        {<}{\textless}{1}
        {>}{\textgreater}{1}
        {\{}{{\{}}{1}
        {\}}{{\}}}{1}
}

\begin{figure*}[htbp]
\centering
\begin{myfigbox}[title={Prompt for Memory Construction}]
\begin{lstlisting}
# Role
You are a Multimodal Memory Agent. Synthesize inputs into a high-density log for the current window.

# Context & Profiles
1. Global Memory: 
{GLOBAL_MEMORY}
2. Face Profiles (Mapping `<face_idx>` to identities):
{INPUT_FACES}

# Current Inputs
3. Timestamps: {START_TIME} to {END_TIME}
4. Visual Stream (1 fps):
{INPUT_IMAGE_SEQUENCE}
5. Audio Stream:
{INPUT_AUDIO_STREAM}
6. Text Stream:
{INPUT_TEXT_STREAM}

# Task
1. Visual Analysis: 
   - If Global Memory is empty: Describe full scene setup (location, layout, present individuals).
   - If Global Memory exists: Describe only CHANGES and NEW ACTIONS.
   - Always use `<face_idx>` for people.

2. Audio Analysis: 
   - Identify speakers (`<face_idx>`) and transcribe dialogue explicitly.
   - Note vocal tone and significant environmental sounds.

3. Semantic (Facts):
   - Extract new facts revealed explicitly or implicitly.
   - Target: Entities, preferences, relationships, physical descriptions, and visible text, etc.
   - Constraint: Must be timeless facts. Strictly exclude temporary actions or general world knowledge.

# Output (Strict JSON)
{
  "visual": "...",
  "auditory": "...",
  "semantic_memory": [
    "Fact 1",
    "Fact 2"
  ]
}
\end{lstlisting}
\end{myfigbox}
\caption{Prompt for multimodal memory construction.}
\label{fig:prompt_mem}
\end{figure*}

\begin{figure*}[htbp]
\centering
\begin{myfigbox}[title={Prompt for Response Generation}]
\begin{lstlisting}
# Role
You are a sophisticated Multimodal AI Agent with memory capability.
# Long-Term Retrieved Memories
1. Semantic Memory:
{RETRIEVED_SEMANTIC_MEMORY}
2. Episodic Memory:
{RETRIEVED_EPISODIC_MEMORY}
# Context & Profiles
1. Global Memory:
{GLOBAL_MEMORY}
2. Face Profiles (Mapping `<face_idx>` to identities):
{INPUT_FACES} 
# Current Inputs
3. Timestamps: {START_TIME} to {END_TIME}
4. Visual Stream (1 fps):
{INPUT_IMAGE_SEQUENCE} 
5. Audio Stream:
{INPUT_AUDIO_STREAM}
6. Text Stream:
{INPUT_TEXT_STREAM}
# Output
Based on the retrieved long-term memories and current context, provide a direct response to the input.
\end{lstlisting}
\end{myfigbox}
\caption{Prompt for response generation.}
\label{fig:prompt_gen}
\end{figure*}

\begin{figure*}[htbp]
\centering
\begin{myfigbox}[title={Prompt for Updating the User Profile}]
\begin{lstlisting}
# Role
User Profiling Agent: Update existing individual profiles using new memory logs.
# Data
1. Current Profiles:
{CURRENT_PROFILES}
2. Memory Logs:
{MEMORY_LOG_SEQUENCE}
# Task & Rules
1. Target Only: Update ONLY the `<face_idx>` keys present in "Current Profiles". Do not add new individuals.
2. Sparse Update: Only output keys for individuals who have *new* information. If no update, omit the key.
3. Explicit Facts: Extract only explicitly seen/heard facts:
    - Identity (Name, age, gender)
    - Persona (Preferences, habits, personality traits)
    - Context (Occupation, roles)
4. Keyword Style: Use only comma-separated keywords or short phrases (max 3 words per fact). 
5. No Redundancy: Do not use synonyms or overlapping traits (e.g., choose one between "Energetic" and "Enthusiastic"). 
6. Consolidate: Merge new facts with existing ones. Keep it dense and concise.
# Output (Strict JSON)
{
  "<face_idx>": {
    "name": "Name",
    "demographics": "Age, Gender, etc.",
    "preferences": "Likes/Dislikes"
  }
}
\end{lstlisting}
\end{myfigbox}
\caption{Prompt for updating the user profile ($\mathcal{M}_p$).}
\label{fig:prompt_profile}
\end{figure*}

\begin{figure*}[!ht]
\centering
\begin{myfigbox}[title={Prompt for Hierarchical Consolidation of Global State}]
\begin{lstlisting}
# Role
You are a Memory Consolidation Agent. Compress the input logs into a unified memory block.

# Input
{MEMORY_LOG_SEQUENCE}
# Task
Group continuous events into merged summaries.
1. Consolidation:
    *   Visual: Synthesize details into a summary of key actions and final states.
    *   Audio: Extract core dialogue and significant sounds.
    *   Assistant: Briefly summarize the assistant's actions and responses.
2. Preservation: Retain all `<face_idx>`, critical actions, and key dialogue.
# Output (Strict JSON)
{
  "visual": "Summary of visual events",
  "auditory": "Consolidated audio record for this group.",
  "assistant": "Summary of assistant's actions and responses."
}
\end{lstlisting}
\end{myfigbox}
\caption{Prompt for hierarchical consolidation of the global state ($\mathcal{S}_g$).}
\label{fig:prompt_global}
\end{figure*}